\RequirePackage{fix-cm}

\documentclass[twocolumn]{svjour3}          
\smartqed  
\usepackage{url}
\usepackage{natbib}
\usepackage{graphicx}
\usepackage{times}
\usepackage{epsfig}
\usepackage{graphicx}
\usepackage{tabularx}
\usepackage{amsmath}
\usepackage{amssymb}
\usepackage{comment}
\usepackage{makecell}
\usepackage{pifont}
\usepackage{multirow}
\usepackage{subfigure}
\usepackage[misc]{ifsym}
\usepackage{afterpage}
\usepackage{setspace}
\usepackage[breaklinks=true]{hyperref}
\usepackage{breakcites}
\usepackage{seqsplit}
\usepackage[table]{xcolor}

\usepackage{wrapfig}
\usepackage{hyphenat}
\usepackage{soul,color}
\usepackage{amsopn}
\usepackage{amsmath}
\usepackage{amssymb}
\usepackage{tabularx}
\usepackage{subfigure}

\definecolor{Red}{cmyk}{0,1,1,0}
\definecolor{Green}{cmyk}{1,0,1,0}
\definecolor{Cyan}{cmyk}{1,0,0,0}
\definecolor{Purple}{cmyk}{0.45,0.86,0,0}
\definecolor{Rosolic}{cmyk}{0.00,1.00,0.50,0}
\definecolor{Blue}{cmyk}{1.00,1.00,0.00,0}
\definecolor{BlueViolet}{cmyk}{0.86,0.91,0,0.04}
\definecolor{NavyBlue}{cmyk}{0.94,0.54,0,0}

\begin{document}

\title{W-HMR: Monocular Human Mesh Recovery in World Space with Weak-Supervised Calibration 
}

\author{Wei Yao$^{1}$      \and
        Hongwen Zhang$^{2}$ \and
        Yunlian Sun$^{1}$  \and
        Yebin Liu$^{3}$ \and
        Jinhui Tang$^1$
}



\institute{Wei Yao \at
              \email{wei.yao@njust.edu.cn}
           \and
           Hongwen Zhang \at
              \email{zhanghongwen@bnu.edu.cn}
           \and
           Yunlian Sun \at
              \email{yunlian.sun@njust.edu.cn}
           \and
            Yebin Liu \at
              \email{ liuyebin@mail.tsinghua.edu.cn}
           \and
           Jinhui Tang \at
              \email{jinhuitang@njust.edu.cn}
           \\
           $^1$ School of Computer Science and Engineering, Nanjing University of Science and Technology,
                Nanjing, China
           \\
           $^2$ School of Artificial Intelligence, Beijing Normal University, Beijing, China
           \\
           $^3$ Department of Automation, Tsinghua University, Beijing, China
}
\date{Received: date / Accepted: date}
\maketitle

\begin{abstract}
Previous methods for 3D human motion recovery from monocular images often fall short due to reliance on camera coordinates, leading to inaccuracies in real-world applications. The limited availability and diversity of focal length labels further exacerbate misalignment issues in reconstructed 3D human bodies. To address these challenges, we introduce W-HMR, a weak-supervised calibration method that predicts “reasonable” focal lengths based on body distortion information, eliminating the need for precise focal length labels. Our approach enhances 2D supervision precision and recovery accuracy. Additionally, we present the OrientCorrect module, which corrects body orientation for plausible reconstructions in world space, avoiding the error accumulation associated with inaccurate camera rotation predictions. Our contributions include a novel weak-supervised camera calibration technique, an effective orientation correction module, and a decoupling strategy that significantly improves the generalizability and accuracy of human motion recovery in both camera and world coordinates. The robustness of W-HMR is validated through extensive experiments on various datasets, showcasing its superiority over existing methods. Codes and demos have been made available on the project page \href{https://yw0208.github.io/w-hmr/}{\textcolor{magenta}{https://yw0208.github.io/w-hmr/}}.
\end{abstract}



\section{Introduction}
\begin{figure*}[ht]
    \centering
    \includegraphics[width=0.99\textwidth]{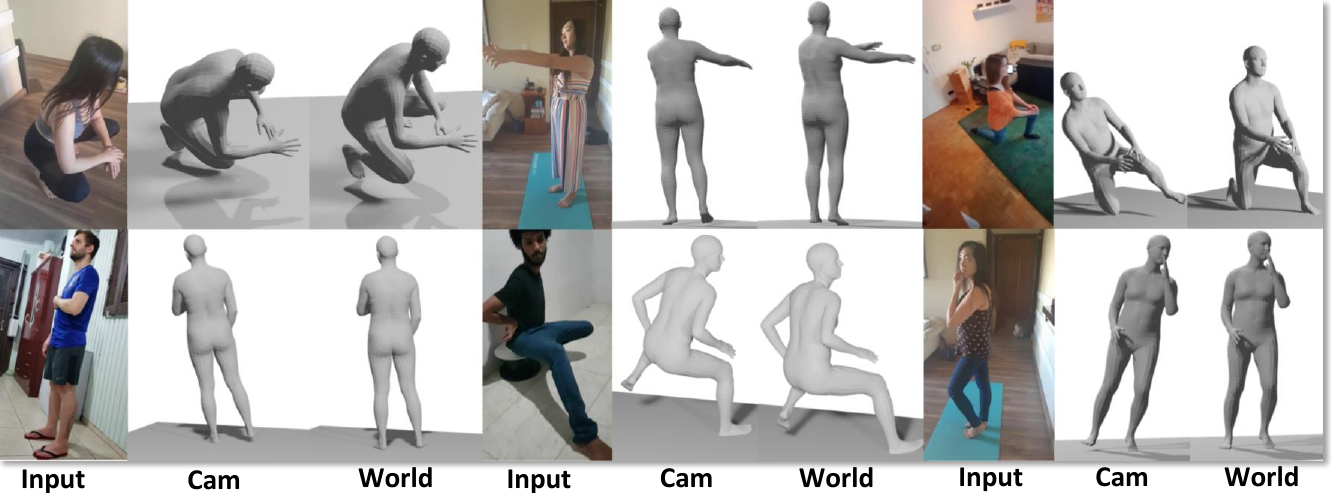}
    \caption{Given \textbf{input} images, we show recovered motion output by traditional methods based on the \textbf{cam}era coordinate and our W-HMR based on the \textbf{world} coordinate. In contrast to traditional methods, W-HMR is capable of rectifying incorrect poses in the camera coordinate and guaranteeing the rationality of poses in the world space.}
    \label{fig:global}
\end{figure*}

{R}{ecovering} 3D human motion from monocular images is crucial in human-centered research, with wide-ranging applications in try-on, animation, and virtual reality. However, there is a noticeable lack of research on global body reconstruction from images captured under complex shooting conditions. Previous methods based on the camera coordinate can only be used in a controlled laboratory environment. When faced with images taken at various strange angles in the wild, previous methods often cannot recover useful results, which can be seen in  Fig.~\ref{fig:global}.

The primary challenge in this task arises from the limited camera information. In the camera parameters, there is an intrinsic parameter focal length that cannot be decoupled. This unknown parameter causes misalignment of the recovered human mesh and the original images. Since most existing models are trained with 2D joint labels, this misalignment due to the unknown focal lengths affects the training and ultimately leads to decreased recovery accuracy. As for extrinsic parameters, camera rotations are essential for coordinate transformation. When the shooting angle makes the world coordinates and camera coordinates inconsistent, the unknown camera rotation limits the application of previous methods based on the camera coordinate. 

\begin{figure}[ht]
    \centering
    \includegraphics[width=0.99\linewidth]{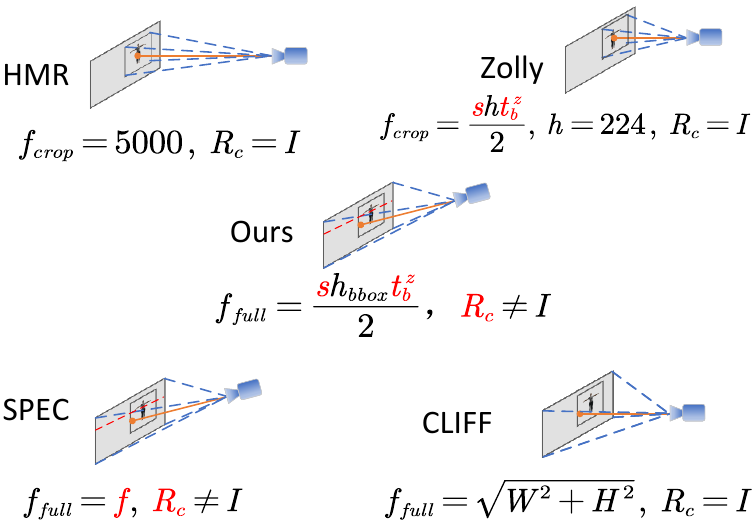}
    \caption{Comparison among the camera models used by the other four models~\citep{hmr,wang2023zolly, spec, cliff} and our W-HMR. $f_{crop}$ and $f_{full}$ denote the focal lengths of the full and cropped images, respectively. 
    {$R_c$ is the camera rotation matrix, and $I$ is the identity matrix. $s$ means scale. $h_{bbox}$ and $h$ are the heights of the bounding box. But $h_{bbox}$ is dynamic, and $h$ is a constant value. $t_{b}^{z}$ refers to the vertical distance from the human body to the plane where the camera is located.}
    The red items mean trainable, i.e., predicted by neural networks.}
    \label{fig:focal}
    \vspace{-2mm}
\end{figure}

Fig.~\ref{fig:focal} illustrates the treatment of the camera in existing models, emphasizing the need for improved methods. The focal length either involves simple manual definitions as seen in HMR~\citep{hmr} and CLIFF~\citep{cliff}, or network predictions as utilized by SPEC~\citep{spec} and Zolly~\citep{wang2023zolly}.
{As for camera rotation, most models~\citep{zhou2010parametric,guan2009estimating, pymaf-x, danet, spin, vibe, kocabas2021pare,lin2021mesh} do not consider it and recover body in the camera coordinate, except for SPEC.}
SPEC, including recent work~\citep{shin2024wham, zhang2023real}, uses a neural network to predict camera rotation to assist human mesh recovery in the world coordinate.
Even based on the most comprehensive camera model, SPEC still has limitations due to its inherent flaws. Using predicted camera rotation as input to inference for world coordinate pose prediction leads to error accumulation, limiting recovery accuracy and reducing model generalizability. A piece of evidence is that SPEC is poor in the camera coordinate compared to its advantage in world space (see Tab.~\ref{tab:main}). To address the above defects, we propose the following two technical contributions.

First, traditional methods such as SPEC predict focal length based on background environment information. This kind of method can predict a coarse focal length to alleviate the problem of misalignment. However, it cannot obtain subtle distortions of body parts and cannot meet the needs of human-centric tasks. In addition, focal length labels are difficult to obtain and not diverse enough to meet the needs of model training. In view of the coarseness of predicted focal lengths and the shortage of labels, we propose a weak-supervised camera calibration method. Specifically, we use more refined mesh-image alignment features to predict focal length. In addition, we use the projection relationship between 3D and 2D joints to supervise the training of our focal length predictor. We let the focal length predictor learn how to predict a suitable focal length to project the predicted 3D joints onto the image and match the 2D joint positions. Compared with traditional methods, our method can capture more detailed misalignment information and is more suitable for human-centric tasks. With our method, 2D supervision becomes more precise and the recovery accuracy is improved.

Second, we adopt a two-stage decoupling method to predict body pose in the world coordinate. We decouple the camera parameters into intrinsics and extrinsics, and the human pose into global body orientation and local joint rotations. In the first stage, we only consider the intrinsic parameter and joint rotations to achieve the best recovery performance in the camera coordinate. Then, we propose a new OrientCorrect module that uses pseudo camera rotation, global information, local joint rotations, and human features to predict global body orientation. Compared with previous methods, the independent OrientCorrect does not harm the existing accuracy of the model. Our method mitigates the impact of incorrect camera rotation on recovery accuracy and achieves higher accuracy in different coordinates simultaneously.

In summary, we decouple the recovery of the global body into three parts: camera calibration, local body reconstruction, and global body orientation correction. This approach prevents errors in camera parameter prediction from accumulating during model training and inference. Our contribution can be grouped into the following three points:

\begin{itemize}
    \item[$\bullet$] We present the first weak-supervised camera calibration method focus on the human recovery task. We achieve reasonable focal length prediction without focal length labels but rely on information about the human body distortion. Better mesh-image alignment makes 2D supervision more precise and effectively improves recovery accuracy. Our method even outperforms the latest supervised method like Zolly~\citep{wang2023zolly}.
    \item[$\bullet$] We design a practical orientation correction module. The module solves the issue that previous methods cannot generate reasonable human poses in world space. This method is simple and effective and can be transplanted to other models without destroying the original structure and performance.
    \item[$\bullet$] Our decoupling idea successfully mitigates the negative impact of incorrect camera parameter estimation. As a result, our model can achieve remarkable reconstruction results in both coordinate systems simultaneously, which is impossible with past models.
\end{itemize}

\section{Related Work}
\subsection{Camera Model in Human Recovery} To visualize the reconstruction results and utilize 2D labels, 3D human recovery models require a camera model. The classical HMR~\citep{hmr} employed a weak-perspective camera model, which assumes a considerable focal length of 5000 pixels to mitigate human distortion. This assumption is suitable for images captured with a long focal length when the human subject is distant from the camera. Due to its simplicity, the weak-perspective camera model has been widely adopted in subsequent works~\citep{pymaf,pymaf-x,kocabas2021pare,humans4d,vibe}.

To our knowledge, Kissos et al.~\citep{kissos2020beyond} was the first to enhance the weak-perspective camera model in pose estimation. They assumed a constant camera Field of View (FoV) of 55 degrees instead of a fixed focal length. Later, CLIFF~\citep{cliff} also adopted the same camera model. SPEC~\citep{spec} took a different approach by constructing a new dataset and training a camera parameter prediction model named CamCalib. The latest method, Zolly~\citep{wang2023zolly}, indirectly predicts the focal length through a well-designed translation prediction module, albeit partially relying on focal length labels and only predicting the focal length of cropped images.

Regarding camera rotation, it was usually overlooked in previous works. SPEC was the pioneering work to advocate for the body's recovery in the world coordinate for postural plausibility. It relied on CamCalib to predict camera parameters and used these parameters as model inputs to directly predict the global body. However, its one-step approach led to error accumulation, constraining reconstruction accuracy and application scope. In contrast, our decoupling of this process effectively addresses these issues.

\subsection{Regression-based Method}In the 3D human body reconstruction domain, methods can be broadly classified into regression-based and optimization-based approaches. Optimization-based methods are beyond the scope of this paper and can be explored in the literature review~\citep{tian2022survey}. Regression-based models can be further categorized into model-based and model-free approaches based on the output type. Model-based models refer to those that output template parameters, with the majority of current models~\citep{hmr, humans4d, pymaf, vibe, kocabas2021pare,li2023mili} still relying on the SMPL algorithm~\citep{loper2015smpl}, and a few~\citep{pons2015dyna, zuffi2015stitched, hund, thundr} based on SCAPE~\citep{anguelov2005scape}, GHUM\&GHUML~\citep{xu2020ghum}, and SMPL-X~\citep{smplx}. These models leveraged the prior knowledge embedded in parameter models but were limited to reconstructing finer motion details due to constraints in degrees of freedom.

In contrast, model-free models in this field output voxels~\citep{varol2018bodynet, zheng2019deephuman}, UV position maps~\citep{zhang2020object, shetty2023pliks} and 3D vertex coordinates~\citep{lin2021mesh,I2l-meshnet,metro,choi2020pose2mesh}. Here, we mainly focus on methods that output 3D vertex coordinates. Apart from methods that output voxels, all other types of output would be converted to 3D vertices and then transformed into meshes based on predefined triangle faces. One classic model-free method, Graphormer~\citep{lin2021mesh}, designed a regression module based on transformer~\citep{vit}, which directly outputs 431 3D sparse vertices. The advantage of model-free models is that they can recover more details than model-based models. However, their limitations lie in their inability to effectively utilize prior knowledge of the human body, and their models tend to have larger sizes. To overcome the limitations of both model-based and model-free approaches, we propose a hybrid method where the model-based regression module provides a strong initialization, and the model-free module further refines the results. This approach aims to achieve optimal mesh precision with relatively fewer parameters, bridging the gap between model-based and model-free methods.

\begin{figure*}[ht]
    \centering
    \includegraphics[width=0.99\textwidth]{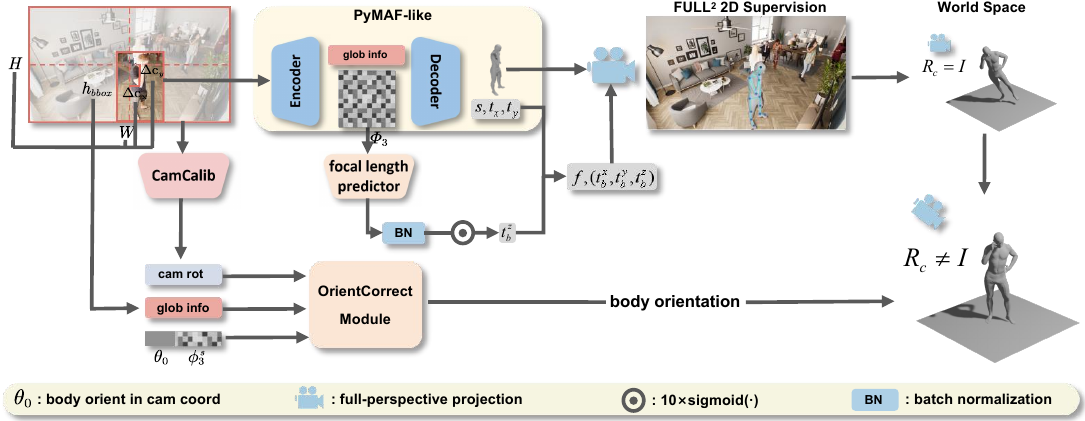}
    \caption{Pipeline of W-HMR. A PyMAF-like backbone is adopted to extract features and predict the recovery results in the camera coordinate.
    {“glob info" refers to global information consisting of five elements. $(\varDelta c_x,\,\varDelta c_y)$ are offsets of the bounding box center relative to the center of the original image. FULL$^2$ 2D supervision refers to 2D joint supervision on \textbf{full} images, which is based on \textbf{full}-perspective camera model. The CamClib takes the whole image as input and outputs camera rotation matrices. The OrientCorrect takes three vectors as input and outputs body orientation in the world coordinate. $\{s,\,t_x,\,t_y\}$ are scales and translations in the camera coordinate. $f$ is focal length and $(t_{b}^{x},\,t_{b}^{y},\,t_{b}^{z})$ is the translation to camera in the world coordinate.}
    Note $\varPhi_3$ and $\phi_3^s$ are features extracted by the PyMAF-like backbone. If you are confused about our feature extraction, please refer to Sec.~\ref{sec:implement} for our network details.}
    \label{fig:model}
\end{figure*}

\section{Method}

The overview of W-HMR is shown in Fig.~\ref{fig:model}. The input image undergoes a PyMAF-like model to extract features and obtain the recovered human motion in the camera coordinates. We design a focal length predictor to predict the focal length to implement a full-perspective projection on the full image for 2D joint supervision. In addition, we propose the OrientCorrect module to correct the body orientation to keep the pose reasonable in the world space. Our method is introduced in five sections. We first present some prior knowledge in Sec.~\ref{sec:Preliminary}. This explains why we need to use the full perspective camera model and the full image information. In Sec.~\ref{sec:focal}, we focus on how we predict the focal length and how we use the predicted focal length for supervision. In Sec.~\ref{sec:orient}, we describe how OrientCorrect works, another essential module. In Sec.~\ref{sec:train}, we describe how to stabilize model training without focal length labels. Finally, in Sec.~\ref{sec:loss}, we add some introduction to more loss functions. To fully demonstrate the effectiveness of our method, we experiment with various network structures. Specific details can be found in Sec.~\ref{sec:implement}.

\subsection{Preliminary}
\label{sec:Preliminary}
\textbf{Perspective Camera Model} In the perspective camera model, camera parameters are divided into intrinsic and extrinsic parameters. The intrinsic parameters $K\in \mathbb{R} ^{3\times 3}$ are generally represented by a matrix as follows

\begin{equation}
    K=\left[ \begin{matrix}
	f_x&		0&		c_x\\
	0&		f_y&		c_y\\
	0&		0&		1\\
\end{matrix} \right] 
\end{equation}

Here, $f_x$ and $f_y$ represent the focal lengths in the x-axis and y-axis directions, respectively. $(c_x,\,c_y)$ is the principal point coordinate. Considering common situations, researchers often assume $f_x=f_y$ and $(c_x,\,c_y)$ is the center of the image. 
{The extrinsic parameters include a camera rotation matrix $R_c\in \mathbb{R} ^{3\times 3}$ and a camera translation vector $t_c\in \mathbb{R} ^{3}$ (camera's position in the world coordinate).}
Most methods based on the camera coordinate generally assume $R_c=I$ and $t_c=(0,\,0,\,0)$. Therefore, we can get $R_b=R_c^{-1}R_b^c=R_b^c$, where $R_b$ is the body orientation in the world coordinate (global orientation), and $R_b^c$ is the body orientation in the camera coordinate (local orientation). This is why conventional models often produce unrealistic human body poses in the real world when dealing with images without known camera parameters.

When we have a 3D point $P=(X,\,Y,\,Z)$ (3D joint or mesh vertex) and want to project it onto the imaging plane to get a 2D point $p=(x,\,y,\,z)$, we calculate following the formula $p^T=K(P+t_b)^T$, where $t_b=(t_{b}^{x},\, t_{b}^{y},\,t_{b}^{z})$ represents the translation of the body relative to the camera. The projection point $p$ is converted into a perspective projection point $p_p=(x_p,\,y_p)$ according to

\begin{equation}
    \left[ \begin{array}{c}
	x_p\\
	y_p\\
\end{array} \right] =\left[ \begin{array}{c}
	\frac{x}{z}\\
	\frac{y}{z}\\
\end{array} \right] =\left[ \begin{array}{c}
	\frac{fX+c_x\left( Z+t_{b}^{x} \right)}{Z+t_{b}^{z}}\\
	\frac{fY+c_y\left( Z+t_{b}^{y} \right)}{Z+t_{b}^{z}}\\
\end{array} \right] 
\label{eq:perspective}
\end{equation}
The whole process is how a 3D object is imaged on a 2D plane based on the perspective camera model. You may notice from Equ.~\ref{eq:perspective} that when $f$ is set very large, the effect of $Z$ would become negligible, and the perspective effect would disappear. The consequence is that the human motion does not align well with the original image, making 2D supervision in error and ultimately affecting recovery accuracy.

\textbf{Translation and Scale} Following ~\citep{hmr}, we continue to utilize SMPL parameters $\varTheta$ as our template. The SMPL parameters encompass pose $\boldsymbol{\theta } \in \mathbb{R} ^{24\times3}$, shape $\boldsymbol{\beta} \in \mathbb{R} ^{10}$, where $\boldsymbol{\theta }_0$ is the body orientation. Specifically, pose $\boldsymbol{\theta }$ and shape $\boldsymbol{\beta }$ can be inputted into a pre-defined function $M\left( \boldsymbol{\theta } ,\boldsymbol{\beta } \right) \in \mathbb{R} ^{N\times 3}$ to obtain $N$ vertex coordinates, i.e., $(X,Y,Z)$ we talked about before. The mesh derived from these vertices is the reconstructed human mesh. To project it onto the image, the model outputs three additional parameters, i.e., translation $(t_x,t_y)$, and scale $\boldsymbol{s}$.  As for $\boldsymbol{s}$ and $(t_x,t_y)$, they determine the size and position of the human meshes within the cropped image. Based on the usual assumptions, we assume a human body range of 2 meters, which establishes the relationship between $t_{b}^{z}$ and $f$ as

\begin{equation}
    \frac{t_{b}^{z}}{2}=\frac{f}{sh_{bbox}}
    \label{fTz}
\end{equation}
where $h_{bbox}$ refers to the height of the bounding box. 
{This formula means that the 2-meter body range is imaged on the image, and the range becomes scale $s$ multiplied by the height of the bounding box $h_{bbox}$. Scale is why this formula is general. The scale is predicted by a neural network, which represents the size of the human in the bounding box. Therefore, it is dynamically adaptive, making this formula valid in most cases. The evidence is qualitative results in our paper. Our recovered mesh can be reprojected to overlap well with the human area. In addition, much work has adopted this assumption and this formula since it was proposed in 2018~\citep{hmr}.} 
For a more detailed reasoning process, please refer to \citep{cliff} or \citep{kissos2020beyond}. Our goal is to leverage the relationship between $t_{b}^{z}$ and $f$ to indirectly predict $f$ by getting $t_{b}^{z}$, thus achieving full perspective. 

\subsection{Weak-Supervised Camera Calibration\label{sec:focal}}
Focal length prediction is essentially an ill-posed problem. From Sec.~\ref{sec:Preliminary}, we can see too many influencing factors in the process, from 3D objects to 2D images. Predicting the actual focal length through a neural network is almost impossible. So, we propose using human body distortion information to predict a “reasonable" focal length for full-perspective projection.

\begin{figure}[h]
    \centering
    \includegraphics[width=0.99\linewidth]{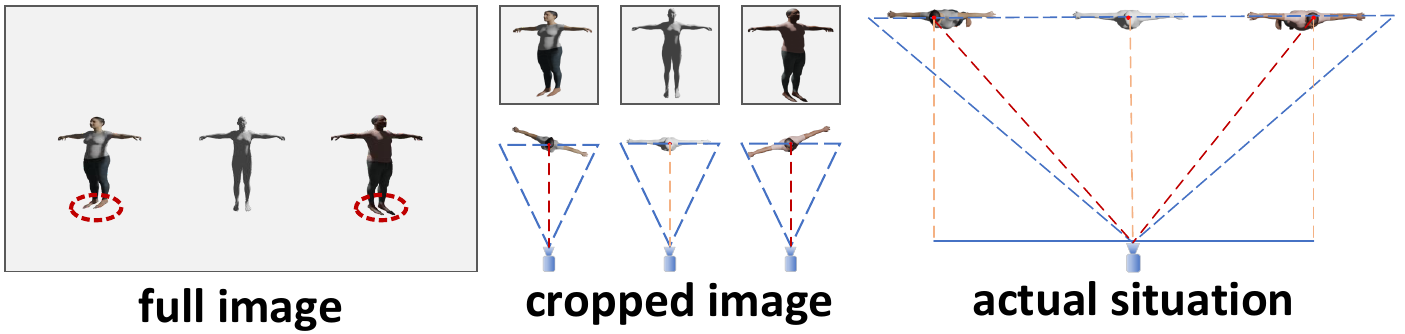}
    \caption{That image on the left is the original image. The middle shows the cropped images used in the traditional method at the top. The predicted focal length, body orientation, and shooting angle in traditional methods are shown at the bottom of the middle image. The image on the right shows the actual shot, illustrating that cropping leads to the loss of necessary information. This loss of necessary information ultimately affects the reconstruction results.}
    \label{fig:whole}
\end{figure}

\textbf{Camera Calibration} The focal length predictor takes the final feature map $\varPhi_3$ as input and output a single value $t_{b}^{z}$. 
{Please refer to appendices for how to extract features $\varPhi_3$.}
To constrain the range of the predicted $t_{b}^{z}$~\citep{wang2023zolly}, we employ a sigmoid function and scale it by 10. This adjustment is necessary because distortion becomes irrelevant when a person stands 10 meters from the camera. We add batch normalization to avoid gradient vanishing or exploding caused by the sigmoid function. The focal length can be calculated according to $f=\frac{sh_{bbox}t_{b}^{z}}{2}$, where the height of bounding box $h_{bbox}$ is known and scale $s$ is output by the PyMAF-like regression module. From Equ.~\ref{eq:perspective}, to implement the projection, we also need to know the translation of the body to the camera. $t_{b}^{z}$ has already been generated, so we need further to compute $(t_b^x,t_b^y)$. Following \citep{kissos2020beyond}, $(t_b^x,t_b^y)=(t_x,t_y)$ when projecting in cropped image. But in the full image, the conversion functions for $(t_x,t_y)$ to $(t_x^b,t_y^b)$ are summarized below

\begin{equation}
    \left[ \begin{array}{c}
	t_{b}^{x}\\
	t_{b}^{y}\\
\end{array} \right] =\left[ \begin{array}{c}
	t_{x}\\
	t_{y}\\
\end{array} \right] +{{\left( 2\left[ \begin{array}{c}
	c^{x}_{bbox}\\
	c^{y}_{bbox}\\
\end{array} \right] -\left[ \begin{array}{c}
	W\\
	H\\
\end{array} \right] \right)}\Bigg/{sh_{bbox}}}
\end{equation}
where $(c^{x}_{bbox},\,c^{y}_{bbox})$ is the center of the bounding box and $(W,H)$ is the width and height of the full image. Projection is important in human recovery because we need to use rich 2D joint labels to enhance supervision and improve model performance. By the way, we also explain why we keep emphasizing the full image here. For example, as shown in Fig.~\ref{fig:whole}, the distortion of stretching to left and right in red circles can only be achieved by projecting on the full image rather than solely on cropped images. That's why we emphasize the implementation of \textbf{full} perspective projection in the \textbf{full} image. Moreover, the actual body orientation cannot be correctly reflected in the cropped image when the human is not in the center of the original image. Therefore, full-image information, i.e., global information in our method, is also an important part of our model input.

\textbf{Weak Supervision} Our main losses can be categorized into joint loss, vertex loss, and parameter loss. In this section, we only introduce the joint loss. Please refer to Sec.~\ref{sec:loss} for information on the others. The joint loss can be calculated using the following function:
\begin{equation}
    \begin{split}
        \mathcal{L}_{joint}&=\lambda _{1}MSE\left( J_{3D},\hat{J}_{3D} \right) \\ &+\lambda _{2}MSE\left( J_{2D},\hat{J}_{2D} \right) \\
        &+\lambda _{3}MSE\left( J_{2D}^{full},\hat{J}_{2D}^{full} \right)\cdot\left( \frac{W}{w_{bbox}},\frac{H}{h_{bbox}} \right)
    \end{split}
    \label{eq:loss}
\end{equation}
where $MSE(\cdot)$ represents mean square error. The $\land$ here denotes ground truth. The $J_{3D}$ denotes 3D joints, which are regressed from mesh vertices\citep{hmr}. $J_{2D}$ signifies normalized 2D joint coordinate in the cropped image and ${J}_{2D}^{full}$ is normalized 2D joint coordinate in the full image. Note that 2D joints are the projected 3D joints on images.
{$J_{2D}$ and $J_{2D}^{full}$ are the same things but based on different coordinates (cropped image vs full image).} 
The last term in the Equ.~\ref{eq:loss} is the key of our FULL$^2$ 2D supervision. It is worth mentioning that $\left( \frac{W}{w_{bbox}},\frac{H}{h_{bbox}} \right)$ is used to dynamically balance the weights and prevent the FULL$^2$ 2D loss from being too small, which could lead to ineffective supervision.

{Weak supervision is reflected in this FULL$^2$ 2D supervision. We do not use any real focal length label to supervise the training of the focal length predictor. From Sec.~\ref{sec:Preliminary}, we know that the focal length is involved in the 3D to 2D projection. So, by having 2D and 3D joint labels, we can let the predictor learn how to generate a correct focal length for 3D to 2D joint projection. After all, all this effort to make the projection more precise is so that the model can be trained more accurately. By the way, rich joint labels are easier to collect for model training than rare focal length labels.}

\subsection{Orientation Correction}
\label{sec:orient}
Our goal is to recover human motion in world space. Weak-Supervised camera calibration can only help the model to get the optimal mesh accuracy in the camera coordinate. After that, we must consider keeping the human pose reasonable in the world coordinate without destroying the original accuracy. In short, we need to predict the body orientation in the world coordinate to replace the body orientation predicted in the camera coordinate.

We propose the OrientCorrect for predicting body orientation in the world coordinate. This module's input comprises the camera rotation $R_c$ predicted by CamCalib, spatial body features $\phi_3^s$, global information, and the body orientation $\theta_0$ in the camera coordinate. The $\theta_0$ is predicted by the PyMAF-like module. The output of OrientCorrect is the human body orientation in the world coordinate. The specific procedure can be expressed as follows:
\begin{equation}
    R_{b}=R_{b}^{c} + FC\left( \oplus \left( \phi _{3}^{s},\, glob\_info,\, R_c,\, R_{b}^{c} \right) \right)
\end{equation}
where $\oplus$ represents concatenation, 
{$FC(\cdot)$ are only couples of FC layers, $R_c$ is the camera rotation output by CamClib, $R_{b}^{c}$ is the predicted body orientation $\theta_0$, and $R_{b}$ is the final predicted body orientation in the world coordinate. Under the supervision of true body orientation in the world coordinate, $FC(\cdot)$ learns to correct the $R_{b}^{c}$ to $R_{b}$. No additional operation is required to ensure the output rotation matrix $R_{b}$ is valid.}
To avoid distress, we explain why the operation of rotation matrices here does not use multiplication but rather direct addition. The network of OrientCorrect is similar to the classic iterative regressor used in HMR~\citep{hmr}. A similar design, instead of multiplication, has been employed in many previous works~\citep{cliff,kocabas2021pare,pymaf,wang2023refit} to predict rotations. The global information 
\begin{equation}
    glob\_info=\left\{ \frac{\varDelta c_x}{\tilde{f}}, \frac{\varDelta c_y}{\tilde{f}}, \frac{h_{bbox}}{\tilde{f}}, \frac{W}{\tilde{f}}, \frac{H}{\tilde{f}}\,\, \right\}
\end{equation}
is important auxiliary information for predicting body orientation~\citep{cliff}. As for $\tilde{f}=\sqrt{W^2+H^2}$ in global information, it is just for normalization here and does not hold any geometric meanings, so we do not use the predicted focal length here. The OrientCorrect is a completely separate module. So, it not only makes full use of what the model has learned in the camera coordinate but also avoids destroying the body feature space that the model has constructed. Additionally, this method can be easily applied to other models based on the camera coordinate without reducing the original precision, shown in Tab.~\ref{tab:orientcorrect}.

\begin{figure*}[ht]
    \centering
    \includegraphics[width=0.99\linewidth]{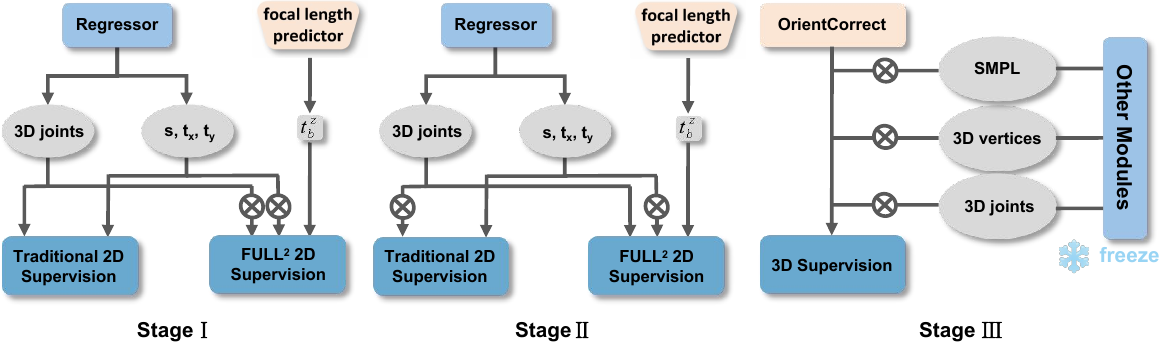}
    \caption{This figure depicts our model training paradigm, illustrating the detachment strategy at different stages to achieve stable model training. The symbol $\otimes$ represents detachment. Traditional 2D supervision refers to projecting 3D joints onto the cropped image to get 2D joints based on the weak-perspective camera model, just as traditional models did. The “freeze" means other modules are frozen in the third stage, and only OrientCorrect is updated.}
    \label{fig:train}
\end{figure*}

\subsection{Training Paradigm}
\label{sec:train}
As shown in Fig.~\ref{fig:train}, our training paradigm is divided into three stages. In stage \uppercase\expandafter{\romannumeral1}, we mainly train the regressors and focal predictors of the model to give them initial predictive capabilities. In stage \uppercase\expandafter{\romannumeral2}, we use FULL$^2$ 2D supervision to improve accuracy further. In stage \uppercase\expandafter{\romannumeral3}, we train OrientCorrect alone to give W-HMR the ability to recover humans in world space. The following is a detailed description of each stage.

In the first stage, after obtaining the 3D joints, we need to project them onto both the cropped image and the original image to obtain $J_{2D}$ and $J_{2D}^{full}$, respectively. 
$J_{2D}$ still adheres to the traditional weak-perspective camera model, while $J_{2D}^{full}$ uses the full-perspective camera model. During the FULL$^2$ 2D supervision, we detach the gradients of the 3D joints, scale $s$, and translation $(t_x,\,t_y)$ to prevent incorrect backpropagation. This precaution is necessary because, in the first stage, the focal length predictor cannot predict a reasonable focal length. This would lead to a completely erroneous 2D joint loss, affecting the regressors to learn to reconstruct the human body. Moreover, to stabilize the training of the focal length predictor, we introduce an additional loss in the first stage, i.e., 
\begin{equation}
    \mathcal{L}_{f}=MSE\left( f,\tilde{f} \right)
\end{equation}
where $\tilde{f}=\sqrt{W^2+H^2}$ serves as pseudo focal length labels to ensure focal length predictor outputs coarse but relatively reasonable results. Note it is only applied in stage \uppercase\expandafter{\romannumeral1}.

Since the model has already developed initial capabilities in predicting joints and focal lengths in the first stage, we can directly apply FULL$^2$ 2D supervision without concern about training collapse. We continue to detach $(s,\,t_x,\,t_y)$ from the full-perspective projection. Because the perspective effect does not impact the model's learning of scale and translation. In the second stage, we detach the gradients of the 3D joints from the weak-perspective projection to prevent erroneous backpropagation of the weak-perspective 2D joint loss on cropped images.

Training is conducted entirely in the camera coordinate in the first two stages. In the third stage, we need to equip the model to reconstruct the human body in the world coordinate. The pre-trained regressors, focal length predictor, and other modules are frozen. We only need to train the OrientCorrect, composed of three FC layers. We exclusively use the 3D joint, mesh vertices, and SMPL parameters in the world coordinate for supervision.

\subsection{Other Losses}
\label{sec:loss}
When we employ a hybrid regression module, we obtain 431 sparse vertices, 1723 coarse vertices, and 6840 complete vertices from the vertex regressor. Therefore, we design a triple vertex loss function, denoted as 
\begin{equation}
    \begin{split}
        \mathcal{L}_{vertex}&=\lambda _4L1\left( V_{sparse},\hat{V}_{sparse} \right) \\ &+\lambda _5L1\left( V_{coarse},\hat{V}_{coarse} \right) \\
        &+\lambda _{6}L1\left( V_{complete},\hat{V}_{complete} \right)
    \end{split}
\end{equation}
where $L1(\cdot)$ is L1 loss. We also directly supervise the obtained SMPL parameters by calculating the function as 
\begin{equation}
    \mathcal{L}_{SMPL}=\lambda _{7}MSE\left( \beta ,\hat{\beta} \right) +\lambda _{8}MSE\left( \theta ,\hat{\theta} \right)
\end{equation}
where $\beta$ is shape parameters in SMPL and $\theta$ is pose parameters represented as rotation matrices.

To enhance the model's ability to focus on human regions and extract refined body information, we include two additional sub-tasks for auxiliary supervision, which are inspired by \citep{pymaf}. 

First, let's delve into the IUV map, which is integral to the IUV loss. The IUV map, as defined in \citep{densepose}, uses “I" to indicate the body part to which a pixel belongs and “UV" to represent the 2D coordinates on the surface of the body part. This mapping establishes dense correspondences from 3D mesh to 2D plane. We employ three 2D convolution layers to predict the IUV map based on the feature map $\varPhi _3$ and render the ground truth online. The loss function is defined as

\begin{equation}
    \begin{split}
        \mathcal{L} _{IUV}&=\lambda _9CrossEntropy\left( P,\hat{P} \right) \\ &+\lambda _{10}SmoothL1\left( \hat{P}\odot U,\hat{P}\odot \hat{U} \right) \\ &+\lambda _{11}SmoothL1\left( \hat{P}\odot V,\hat{P}\odot \hat{V} \right) 
    \end{split}
\end{equation}
where $\odot$ means masking, which indicates only the foreground pixels are applied. 

The IUV loss focuses solely on supervising the 2D information, neglecting the depth information essential for capturing distorted details. To address this limitation, we introduce an additional depth map loss. We generate ground truth depth maps by rendering them online. A 2D convolution is used to predict the depth map based on the feature map $\varPhi _3$. The loss is computed as follows:

\begin{equation}
    \mathcal{L}_{depth}=\lambda_{12}L1_{smooth}\left( \frac{1}{D},\frac{1}{\hat{D}} \right) 
\end{equation}
where we use the reciprocal of the depth for normalization.

\section{Experiment}

\subsection{Implementation Details}
\label{sec:implement}

We train three models in total for evaluation. Two of them utilize a hybrid regression module, using 431 sparse vertices for downsampling. The distinction between these two models lies in using different backbones, including ResNet50~\citep{resnet} and VitPose-B~\citep{xu2022vitpose}. Additionally, we train an extra model using a pure parameter regression module, employing 67 mocap markers for downsampling, with the backbone being VitPose-B. For training, we utilize datasets such as Human 3.6m~\citep{human36m}, 3DPW~\citep{3dpw}, MPII3D~\citep{MPII3d}, MPII~\citep{mpii}, AGORA~\citep{agora}, COCO~\citep{coco}, SPEC-SYN~\citep{spec}, and HuMMan~\citep{cai2022humman}. Some datasets are augmented with pseudo-labels generated by CamCalib, EFT~\citep{eft}, or CLIFF~\citep{cliff}. Please refer to appendices for detailed information about the datasets used. Our model can simultaneously provide results in camera and world coordinates. But if results in the world coordinate are not needed or camera rotation is known, the CamCalib and OrientCorrect can be discarded. The focal length predictor can also be discarded if mesh projection is not required.

\begin{table*}[ht]

\caption{Comparison with SOTA methods on distorted datasets in the camera coordinate. Some of the evaluation results on HuMMan~\citep{cai2022humman} and SPEC-MTP~\citep{spec} are sourced from ~\citep{wang2023zolly}, while the evaluation results on AGORA~\citep{agora} are obtained from its official website \href{https://agora-evaluation.is.tuebingen.mpg.de/}{https://agora-evaluation.is.tuebingen.mpg.de/}. Including the following tables, $^*$ represents backbone is ResNet50~\citep{resnet}, $^\dagger$ means VitPose-B~\citep{xu2022vitpose,xu2022vitpose+}, and “P" denotes the pure parameter module. Note we do not perform any fine-tuning on SPEC-MTP.}
\label{tab:main}
\scriptsize%
\centering%
\setlength{\tabcolsep}{2.2mm}{
\begin{tabular}{ccccccccccc}
\hline
\multirow{2}{*}{Model}                             & \multicolumn{4}{c}{AGORA}                                                                                                         & \multicolumn{3}{c}{HUMMAN}                                                                       & \multicolumn{3}{c}{SPEC-MTP}                                                                       \\ \cline{2-11} 
                                                   & NMVE$\downarrow$               & NMJE$\downarrow$               & MVE$\downarrow$                & MPJPE$\downarrow$              & MPJPE$\downarrow$              & PA-MPJPE$\downarrow$           & PVE$\downarrow$                & MPJPE$\downarrow$               & PA-MPJPE$\downarrow$           & PVE$\downarrow$                 \\ \hline
HMR~\citep{hmr}               & 217.0                          & 226.0                          & 173.6                          & 180.5                          & 43.6                           & 30.2                           & 52.6                           & 121.4                           & 73.9                           & 145.6                           \\
HMR-f~\citep{hmr}             & -                              & -                              & -                              & -                              & 43.6                           & 29.9                           & 53.4                           & 123.2                           & 72.7                           & 145.1                           \\
SPEC~\citep{spec}             & 126.8                          & 133.7                          & 106.5                          & 112.3                          & 44.0                           & 31.4                           & 54.2                           & 125.5                           & 76.0                           & 144.6                           \\
CLIFF~\citep{cliff}           & 83.5                           & 89.0                           & 76.0                           & 81.0                           & 42.4                           & 28.6                           & 50.2                           & 115.0                           & 74.3                           & 132.4                           \\
PARE~\citep{kocabas2021pare}  & 167.7                          & 174.0                          & 140.9                          & 146.2                          & 53.2                           & 32.6                           & 65.5                           & 121.6                           & 74.2                           & 143.6                           \\
GraphCMR~\citep{graphcmr}     & -                              & -                              & -                              & -                              & 40.6                           & 29.5                           & 48.4                           & 121.4                           & 76.1                           & 141.6                           \\
FastMETRO~\citep{fastmetro}   & -                              & -                              & -                              & -                              & 38.8                           & 26.3 & 45.5                           & 123.1                           & 75.0                           & 137.0                           \\
PLIKS~\citep{shetty2023pliks} & 71.6                           & 76.1 & 67.3                           & 71.5                           & -                              & -                              & -                              & -                               & -                              & -                               \\
ProPose~\citep{propose}       & 78.8                           & 82.7                           & 70.9                           & 74.4                           & -                              & -                              & -                              & -                               & -                              & -                               \\
HybrIK~\citep{li2021hybrik}   & 81.2                           & 84.6                           & 73.9                           & 77.0                           & -                              & -                              & -                              & -                               & -                              & -                               \\
Hand4Whole~\citep{hand4whole} & 90.2                           & 95.5                           & 84.8                           & 89.8                           & -                              & -                              & -                              & -                               & -                              & -                               \\
PyMAF~\citep{pymaf}                                           & 87.3                           & 92.6                           & 78.6                           & 83.3                           & 42.5 & 28.9                           & 50.4                           & 108.6 &66.1 & 127.5                           \\
Zolly~\citep{wang2023zolly}   & -         & -      & -       & -      & \textbf{32.6} & \textbf{22.3}                           & \textbf{40.0}                           & 114.6 &67.4 & 126.7                           \\\hline
Ours$^*$                                           & 78.1                           & 87.0                           & 70.3                           & 78.3                           & 37.4 & 29.1                           & 40.5                           & \textbf{104.5} & \textbf{61.7} & 118.7                           \\
Ours$^\dagger$                                     & \textbf{68.7} & 77.7                           & \textbf{61.8} & 69.9 & 39.2                           & 30.8                           & 40.4 & 104.9                           & 62.3                           & \textbf{115.2} \\
Ours-P$^\dagger$                                               & 70.4                           & \textbf{75.4}                           & 63.4                           & \textbf{67.9}                           & 36.5                           & 27.1                           & 43.5                           & 108.1                           & 65.7                           & 121.5                           \\ \hline
\end{tabular}}
\end{table*}

We experiment with different backbones and downsampling points to validate our method and explore the optimal model structure. As a result, there are corresponding variations in the sizes of input images and features. We summarize these variations in the appendices to provide an overview of these changes. The training process is conducted with a batch size of 128 and a fixed learning rate of 0.00005 on three RTX 3090 GPUs until convergence, taking approximately one week. Following \citep{pymaf,kocabas2021pare}, we apply standard data augmentation techniques such as rotation, flipping, scaling, translation, noise, and occlusion. As for hyperparameters, $\lambda_1=\lambda_2=\lambda_3=300$, $\lambda_4=\lambda_5=\lambda_{6}=15$, $\lambda_{7}=0.06$, $\lambda_{8}=60$, $\lambda_9=0.03$ and $\lambda_{10}=\lambda_{11}=\lambda_{12}=0.05$. 


\subsection{About Metrics}

We evaluate the reconstruction accuracy using 3D mesh vertices and joints. Specifically, for AGORA, we employ four metrics: MPJPE, MVE, NMJE, and NMVE. MPJPE stands for Mean Per Joint Position Error, which measures the accuracy of joint positions. For AGORA, it calculates the error for the first 24 SMPL joints after aligning with the root joint. Following \citep{pymaf,spin}, we utilize 17 joints for MPI-INF-3D, while the other datasets employ a consistent set of 14 keypoints. MVE, or Mean Vertex Error, measures the error of all SMPL mesh vertices after aligning the root vertices. PVE (Per Vertex Error) is computed without alignment for other datasets except AGORA. These metrics help assess the accuracy of the reconstructed human bodies. NMJE (Normalized Mean Joint Error) and NMVE (Normalized Mean Vertex Error) are obtained by normalizing MPJPE and MVE using the F1 score. They provide an overall evaluation of the method's performance, particularly in handling challenging scenarios. For specific calculation methods, please refer to \citep{agora}.

As for the metrics in the world coordinate, we follow SPEC~\citep{spec} using W-MPJPE, PA-MPJPE, and W-PVE. PA-MPJPE (Procrustes-aligned Mean Per Joint Position Error) exists as a metric that purely evaluates the accuracy of body reconstruction. It removes the effect of unknown camera rotation by Procrustes-aligned and has therefore been adopted by many previous works. The calculation of W-MPJPE and W-PVE is the same as that of MPJPE and PVE. The only difference is that the labels used are adjusted to the ground truth in the world coordinate. We, including SPEC, believe these metrics represent the model's capabilities in real-world applications.

\subsection{Evaluation Results}
We conduct massive experiments on multiple benchmarks in a variety of formats. The evaluation results can be categorized into three main parts. Firstly, we do experiments on the distorted dataset (AGORA, HuMMan, and SPEC-MTP) in camera and world coordinates, respectively, to demonstrate the effectiveness of W-HMR on distorted images. Secondly, we also do rich experiments on traditional benchmarks (3DPW, H36M, and MPI-INF-3D) to compare the past SOTA methods to demonstrate our model's superiority. Finally, we provide rich qualitative results to visualize the performance of W-HMR.

\textbf{Camera Coordinate} In our study, we choose three challenging datasets containing distorted images to showcase the superiority of weak-supervised camera calibration under complex conditions. As depicted in Tab.~\ref{tab:main}, our method surpasses the performance of most existing methods on the challenging synthetic dataset AGORA~\citep{agora}. Each of our models demonstrates unique strengths. The pure regression model achieves higher accuracy in joint localization, and the hybrid regression module excels in capturing mesh details. Compared to baseline~\citep{pymaf}, our improvement is quite substantial. Note the baseline's backbone, training datasets, and evaluation settings are the same as Ours$^*$. Although the joint accuracy does not exhibit significant improvement on the indoor dataset HuMMan~\citep{cai2022humman}, we achieve the closest performance compared to the latest method Zolly in PVE, even surpassing the model-free method FastMETRO~\citep{fastmetro}. The most remarkable performance is observed on the real distorted dataset SPEC-MTP. We significantly elevate the SOTA results on SPEC-MTP to new heights.

\begin{table}[th]

\caption{Comparison with SOTA methods on SPEC-MTP~\citep{spec}. The first six models in the table are all methods that output results in the camera coordinate, so there are two results. The “a" in “a/b" indicates the result calculated directly using the output in the camera coordinate. The “b" is the result calculated after camera rotation predicted by CamCalib has calibrated the output. The latter two models can directly get the results in the world coordinate without calibration, so there is only one result.}
\label{tab:global}
\scriptsize%
\centering%
\setlength{\tabcolsep}{4mm}{
\begin{tabular}{clccc}
\hline
\multicolumn{2}{c}{\multirow{2}{*}{Model}} & \multicolumn{3}{c}{SPEC-MTP}                    \\ \cline{3-5} 
\multicolumn{2}{c}{}                       & W-MPJPE$\downarrow$        & PA-MPJPE$\downarrow$      & W-PVE$\downarrow$          \\ \hline
\multicolumn{2}{c}{GraphCMR}               & 175.1/166.1    & 94.3          & 205.5/197.3    \\
\multicolumn{2}{c}{SPIN}                   & 143.8/143.6    & 79.1          & 165.2/165.3    \\
\multicolumn{2}{c}{PartialHumans}          & 158.9/157.6    & 98.7          & 190.1/188.9    \\
\multicolumn{2}{c}{I2L-MeshNet}            & 167.2/167.0    & 99.2          & 199.0/198.1    \\
\multicolumn{2}{c}{HMR}                    & 142.5/128.8    & 71.8          & 164.6/150.7    \\
\multicolumn{2}{c}{PyMAF}                 & 148.8/134.2    & 66.7         & 166.7/158.7    \\
\multicolumn{2}{c}{Ours-P$^\dagger$}                 & 139.3/130.3    & \textbf{66.6}          & 155.8/147.8    \\ \hline
\multicolumn{2}{c}{SPEC}                   & 124.3          & 71.8          & 147.1          \\
\multicolumn{2}{c}{Ours-P$^\dagger$}                 & \textbf{118.7} & \textbf{66.6} & \textbf{133.9} \\ \hline
\end{tabular}}
\end{table}

\textbf{World Coordinate} We utilize SPEC-MTP~\citep{spec} as the primary evaluation dataset. Due to its only test set, specific fine-tuning is not feasible. It adopts a shooting method different from traditional datasets, resulting in unique rotation and severe distortion. Therefore, it poses a significant challenge and is suitable for demonstrating OrientCorrect's strength. We only evaluate the W-HMR with a pure parameter regression module to ensure fairness. W-HMR easily outperforms existing methods, including SPEC. It is worth noting that the superior performance of W-HMR in the world coordinates is not entirely dependent on higher reconstruction accuracy. Perhaps you noticed two Ours-Ps in Tab.~\ref{tab:global}. This is because W-HMR can generate results in two coordinates simultaneously. By comparing these two sets of results, we can conclude that OrientCorrect is significantly superior to the traditional method of directly using predicted camera rotation for correction. Upon comparing our results calibrated by CamCalib with those corrected by OrientCorrect, it is evident that our module is effectively functioning. As illustrated in Fig.~\ref{fig:global}, the reconstructed human body pose appears implausible in world space when the camera coordinate is not aligned with the world coordinate. However, our OrientCorrect corrects the orientation of the human body and maintains its reasonableness in the world coordinate. Our model can accurately perceive and correct both slight and significant camera rotation. Quantitative and qualitative results demonstrate that our method is simpler and more efficient than SPEC, which will be further substantiated in subsequent ablation studies.

\begin{table}[h]
\caption{Comparison with SOTA Methods on controllable datasets including Human 3.6M~\citep{human36m} and MPI-INF-3D~\citep{MPII3d}.}
\label{tab:h36m}
\scriptsize%
\centering%
\setlength{\tabcolsep}{2.8mm}{
\begin{tabular}{ccccc}
\hline
\multirow{2}{*}{Model} & \multicolumn{2}{c}{H36M}       & \multicolumn{2}{c}{MPI-INF-3D} \\ \cline{2-5} 
                       & MPJPE$\downarrow$         & PA-MPJPE$\downarrow$       & MPJPE$\downarrow$        & PA-MPJPE$\downarrow$        \\ \hline
HMR                    & 88.0          & 56.8           & 124.2        & 89.8            \\
GraphCMR               & -             & 50.1           & -            & -               \\
SPIN                   & 62.5          & 41.1           & 105.2        & 67.5            \\
PyMAF                  & 57.7          & 40.5           & -            & -               \\
FastMETRO              & 53.9 & 37.3           & -            & -               \\
HMMR                   & -             & 56.9           & -            & -               \\
VIBE                   & 78.0          & 53.3           & 96.6         & 64.6            \\
MEVA                   & 76.0          & 53.2           & 96.4         & 65.4            \\
HybrIK         & 54.4          & 34.5           & -         & -            \\
CLIFF                   & 47.1          & 32.7           & -        & -            \\
MPS-Net                & 69.4          & 47.4           & 96.7         & 62.8            \\ 
STAF                & 70.4          & 44.5           & 92.4         & \textbf{58.8}            \\ 
Zolly                & 49.4          & 32.3           & -         & -            \\ \hline
Ours$^*$               & 52.7         & 34.0          &88.3             &62.0                 \\
Ours$^\dagger$         & 51.0         & \textbf{29.8} &\textbf{83.2}              &59.1                 \\
Ours-P$^\dagger$         & \textbf{45.5}         & 30.2 &83.3              &59.8                 \\
\hline
\end{tabular}}
\end{table}

\textbf{Traditional Benchmarks} Following previous methods \citep{pymaf,mpsnet}, we conduct evaluation on 3DPW~\citep{3dpw}, H36M~\citep{human36m}, and MPII-INF-3D~\citep{MPII3d}. Specifically, we evaluate the performance of models trained with and without the 3DPW train set on the 3DPW test set. As shown in Tab.~\ref{tab:3dpw}, we achieve remarkable improvements on this challenging in-the-wild dataset, surpassing previous wor-ks easily. Even compared to the SOTA method Zolly, W-HMR still exhibits significant advancements when trained without the 3DPW train set, which shows W-HMR's robustness. In Tab.~\ref{tab:h36m}, W-HMR's performance is relatively less impressive in the case of controllable datasets, although it remains competitive with these SOTA methods. Controllable datasets refer to datasets obtained under laboratory conditions, typically characterized by limited variations in actors and scenes, which are prone to overfitting. HuMMan in Tab.~\ref{tab:main} is an example of such datasets. So, it becomes difficult for our advantages to be fully demonstrated.

\begin{table}[h]

\caption{Comparison with SOTA Methods on 3DPW~\citep{3dpw}. The column “w/ 3DPW" indicates whether the model has been trained using the 3DPW dataset. We bold both the best results with and without 3DPW.}
\label{tab:3dpw}
\scriptsize%
\centering%
\setlength{\tabcolsep}{3.3mm}{
\begin{tabular}{ccccc}
\hline
\multirow{2}{*}{Model} & \multirow{2}{*}{w/ 3DPW} & \multicolumn{3}{c}{3DPW}                      \\ \cline{3-5} 
                       &                          & MPJPE$\downarrow$         & PA-MPJPE$\downarrow$      & PVE$\downarrow$           \\ \hline
HMR                    & No                       & 130.0         & 76.7          & -             \\
GraphCMR               & No                       & -             & 70.2          & -             \\
SPIN                   & No                       & 96.9          & 59.2          & 116.4         \\
SPEC                   & Yes                      & 96.4          & 52.7          & -             \\
I2L-MeshNet            & No                       & 93.2          & 57.7          & 110.1         \\
Graphomer              & Yes                      & 74.7          & 45.6          & 87.7          \\
FastMETRO              & Yes                      & 73.5          & 44.6          & 84.1          \\
PyMAF                  & Yes                      & 92.8          & 58.9          & 110.1         \\
VIBE                   & No                       & 93.5          & 56.5          & 113.4         \\
VIBE                   & Yes                      & 82.9          & 51.9          & 99.1          \\
SATF                   & No                       & 81.2          & 48.7          & 96.0         \\
STAF                   & Yes                      & 80.6          & 48.0          & 95.3          \\
PARE                   & No                       & 84.1          & 49.3          & 99.4          \\
PARE                   & Yes                      & 79.1          & 46.4          & 94.2          \\
HybrIK                 & No                       & 80.0          & 48.8          & 94.5          \\
HybrIK                 & Yes                      & 74.1          & 45.0          & 86.5   \\
Zolly                 & No                       & 76.2          & 47.9          & 89.8          \\
Zolly                 & Yes                      & 65.0          & \textbf{39.8}          & 76.3
\\\hline
Ours$^*$               & No                       & \textbf{75.6}          & \textbf{46.7}          & \textbf{87.7}           \\
Ours$^\dagger$         & No                       & 85.4          & 54.6          & 96.0          \\
Ours-P$^\dagger$         & No                       & 82.9          & 52.1          & 95.1          \\
Ours$^*$               & Yes                      & 68.5          & 43.2          & 81.6          \\
Ours$^\dagger$         & Yes                      & \textbf{64.6} & 40.5 & \textbf{75.7} \\
Ours-P$^\dagger$         & Yes                      & 67.3 & 41.6 & 79.9 \\ \hline
\end{tabular}}
\end{table}

\begin{figure}[h]
    \centering
    \includegraphics[width=\linewidth]{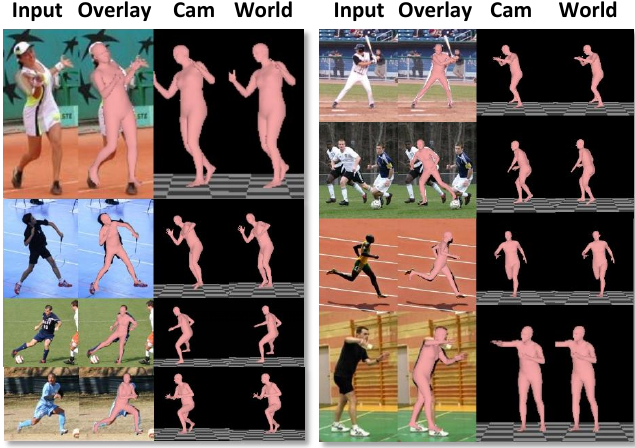}
    \caption{Qualitative results in camera and world coordinates. W-HMR shows excellent reconstruction accuracy and robustness. It can reconstruct various human bodies well in different scenes. Not only does the reconstructed mesh align well with the original image, but even when switching the angle of view, the reconstructed human body can still be seen to fit the target human.}
    \label{fig:lsp}
\end{figure}

\textbf{Qualitative Results} More visualizations are generated to give a more comprehensive reference. Also, we welcome running our codes on your samples. Back to qualitative results, from Fig.~\ref{fig:lsp}, we can see that our model can recover the human body very well when facing various scenes with various characters in multiple poses. We purposely show the results from side angles and in world space. In the past, many methods could also make the reconstructed mesh match the original image perfectly, but it would not look perfect if we changed the visual angle. As shown in Fig.~\ref{fig:lsp}, W-HRM does not suffer from this issue and performs well in every condition. 

\begin{figure*}[ht]
    \centering
    \includegraphics[width=0.99\linewidth]{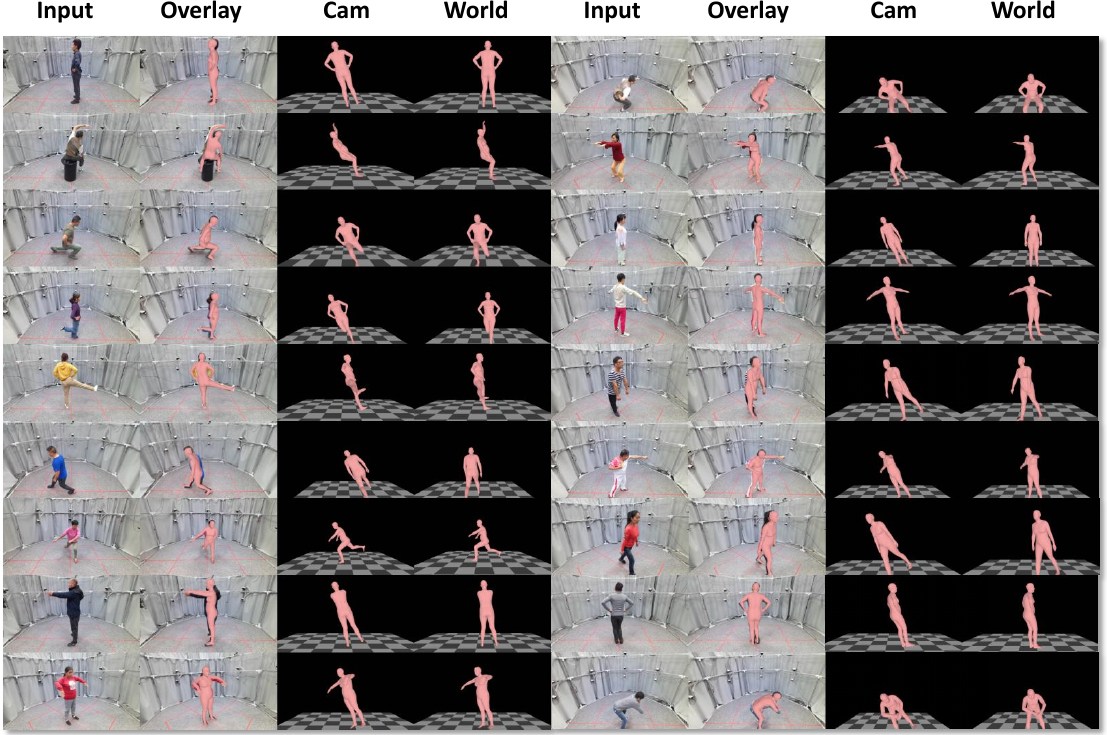}
    \caption{Qualitative results on HuMMan~\citep{cai2022humman}. “Input" refers to input images of the model. “Overlay" means we project the recovered mesh back onto the original images. “Cam" and “World" show the output in camera and world coordinates, respectively.}
    \label{fig:humman}
\end{figure*}

\begin{figure}[h]
    \centering
    \includegraphics[width=\linewidth]{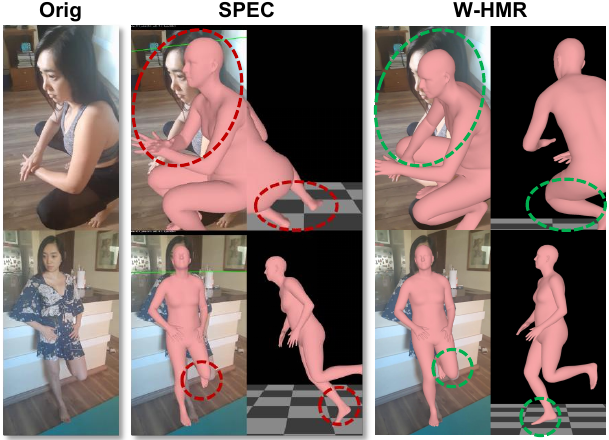}
    \caption{Qualitative comparison with SPEC on SPEC-MTP~\citep{spec}. 
    {Here are some challenging examples. Although imperfect, W-HMR shows better mesh-image alignment and higher orientation accuracy in world space.}
    }
    \label{fig:specmtp}
\end{figure}

Methods based on the camera coordinate often fail to meet the needs of applications in world space. It may not be noticeable from Fig.~\ref{fig:lsp} because they fit the weak-perspective camera model and the assumption of $R_c=I$. But from Fig.~\ref{fig:humman}, the drawbacks of the traditional method are undeniable. The reconstructed human body is entirely lopsided in the world coordinate. Instead, our OrientCorrect module can correct the pose well to achieve a reasonable pose recovery in the world coordinate. This method has many applications, allowing the model to reconstruct a proper human body when facing in-the-wild images. Before W-HMR, there were also methods like SPEC that explored human motion recovery in world space. As a pioneering work, SPEC opened up a new way of thinking for subsequent research, but the performance was not robust enough. In Fig.~\ref{fig:specmtp}, we can see that SPEC is not as accurate as W-HMR in any coordinate, showing that our method cleverly avoids the error accumulation of SPEC and achieves all-around transcendence.

{\textbf{Camera Parameter} There is no such evaluation for inferred camera parameters because W-HMR cannot and does not need to predict true intrinsic and extrinsic parameters. First, focal length prediction is an ill-posed problem and lacks much training data. We choose to leverage mesh-image alignment information to predict pseudo focal lengths. This pseudo focal length may differ from the true focal length but can accurately project the recovered mesh to the human area. After all, the purpose of predicting the focal length is to handle distorted images so that the calculation of 2D joint loss is more accurate, thereby improving model performance. In summary, inferred camera parameters cannot be evaluated quantitatively because they may differ significantly from ground truth. Of course, it can be implicated by improved recovery accuracy and aligned-well qualitative results. Moreover, the module used to predict the camera rotation matrix in W-HMR comes from CamClib in SPEC. Its accuracy has been evaluated in SPEC. The novelty in our work is how we better use this information. With the same CamClib, W-HMR still significantly outperforms SPEC, which is a solid evidence to prove the effectiveness of our method.}

\subsection{Ablation Study}

Our ablation study is primarily divided into three parts. The first part focuses on ablating our model's camera calibration to demonstrate our approach's effectiveness in enhancing reconstruction accuracy. More specifically, we explore the interplay between the focal length predictor and other designs. The second part involves ablating different global orientation recovery methods. We show the limitations of SPEC and demonstrate the robustness of OrientCorrect. In the third part, we do more ablation studies about some functional designs but not our core contribution.

\textbf{Camera Calibration} We select PyMAF\citep{pymaf} as our baseline and consistently use ResNet50 as the backbone throughout our ablation study, with 431 sparse vertices as the downsampling points. The training and evaluation settings are all kept the same. From the Tab.~\ref{tab:ablation}, it is evident that the most significant improvements stem from adding the vertex regressor and global information. Including the vertex regressor enhances the model's ability to reconstruct body mesh details. Moreover, global information contributes independently and aids in camera calibration. A comparison between “V" and “V+F" reveals that solely adding the focal length predictor only has a noticeable impact on vertex accuracy. However, comparing “V+G" and “V+F+G" demonstrates that the presence of global information leads to more substantial improvements when combined with the focal length predictor. This finding further supports our idea that global information is crucial for recovering body orientation, predicting focal lengths, and correcting distortions. In summary, with the camera calibration assistance, W-HMR achieves a decent reconstruction accuracy, which provides a solid foundation for human body recovery in world space.

\begin{table}[h]

\caption{The ablation study on the core block of the W-HMR. The experiments below utilize ResNet50~\citep{resnet} as the backbone. In this table, “V" refers to the vertex regressor. “F" indicates whether the focal length predictor and FULL$^2$ 2D supervision are implemented. “G" signifies the addition of global information during input.}
\label{tab:ablation}
\scriptsize%
\centering%
\setlength{\tabcolsep}{3.1mm}{
\begin{tabular}{ccccccc}
\hline
                    &                     &                     &                      \multicolumn{4}{c}{AGORA}                                                                                                         \\ \cline{4-7} 
\multirow{-2}{*}{V} & \multirow{-2}{*}{F} & \multirow{-2}{*}{G}  & NMVE$\downarrow$                           & NMJE$\downarrow$                           & MVE$\downarrow$                            & MPJPE$\downarrow$                          \\ \hline 
                    &                     &                                          & 87.3                           & 92.6                           & 78.6                           & 83.3                           \\
\rowcolor[HTML]{EFEFEF}
$\surd$             &                     &                                          & 82.4                           & 88.6                           & 75.8                           & 81.5                           \\ 
$\surd$             & $\surd$             &                                          & 81.2                           & 89.3                           & 74.7                           & 82.2                           \\
\rowcolor[HTML]{EFEFEF}
$\surd$             &                     & $\surd$                                  & 80.7                           & 87.7                           & 73.4                           & 79.8                           \\\hline
$\surd$             & $\surd$             & $\surd$                                  & \textbf{79.8}                           & \textbf{87.6} & \textbf{72.6}                           & \textbf{79.7} \\ \hline
\end{tabular}}
\end{table}

\textbf{Orientation Correction} We select two methods for evaluation. The first method is similar to SPEC. SPEC concatenates features and the camera parameters obtained from CamCalib, then inputs them into a regressor to directly recover the human body in the world coordinate. As demonstrated in Tab.~\ref{tab:global_ablation}, the spec-like method significantly outperforms previous methods. However, we attribute this performance to overfitting. The model relies on massive parameters to memorize the global body orientation in the training set of SPEC-SYN, subsequently achieving excellent results on the test set. This is supported by the inferior performance of spec-like methods on SPEC-MTP. This indicates that such a method is prone to overfitting, leading to the misconception that the model can capture environmental information. Worse, this design destroys the model's original learned knowledge, making the overall reconstruction less precise. In contrast, our method offers a more straightforward structure, requires less training data, and exhibits greater robustness. Furthermore, other models based on the camera coordinate can easily adopt our approach to achieve global pose recovery.

\begin{table}[h]

\caption{The ablation study on body orientation correction. “spec-like" refers to using a similar method like SPEC to predict global pose directly. The numbers in brackets are the evaluation results on SPEC-MTP. This is the reason why we do not choose the spec-like method. Its impressive results are based on overfitting and sacrificing generalization.}
\label{tab:global_ablation}
\scriptsize%
\centering%
\setlength{\tabcolsep}{3.5mm}{
\begin{tabular}{clccc}
\hline
\multicolumn{2}{c}{\multirow{2}{*}{Model}} & \multicolumn{3}{c}{SPEC-SYN}                                      \\ \cline{3-5} 
\multicolumn{2}{c}{}                       & W-MPJPE$\downarrow$              & PA-MPJPE$\downarrow$            & W-PVE$\downarrow$                \\ \hline
\multicolumn{2}{c}{GraphCMR}               & 181.7/181.5          & 86.6                & 219.8/218.3          \\
\multicolumn{2}{c}{SPIN}                   & 165.8/161.4          & 79.5                & 194.1/188.0          \\
\multicolumn{2}{c}{PartialHumans}          & 169.3/174.1          & 88.2                & 207.6/210.4          \\
\multicolumn{2}{c}{I2L-MeshNet}            & 169.8/163.3          & 82.0                & 203.2/195.9          \\
\multicolumn{2}{c}{HMR}                    & 128.7/96.4           & 55.9                & 144.2/111.8          \\
\multicolumn{2}{c}{PyMAF}                 & 126.8/88.0                & 48.7                & 136.7/118.0                \\
\multicolumn{2}{c}{Ours-P$^\dagger$}                 & 127.8/106.2                & 46.2                & 137.8/132.6                \\ \hline
\multicolumn{2}{c}{SPEC}                   & 74.9(124.3)          & 54.5(71.8)          & 90.5(147.1)          \\
\multicolumn{2}{c}{Ours-P$^\dagger$(spec-like)}      & \textbf{68.2}(185.0) & \textbf{43.6}(63.3) & \textbf{76.4}(208.7) \\
\multicolumn{2}{c}{Ours-P$^\dagger$}                 & 82.1(\textbf{118.7}) & 46.1(\textbf{66.6}) & 93.3(\textbf{133.9}) \\ \hline
\end{tabular}}
\end{table}

To verify the portability of OrientCorrect, we do related experiments, and the results are displayed in Tab.~\ref{tab:orientcorrect}. To make the experiments more convincing, we choose HMR, a widely adopted model, as the baseline. We design two porting methods. One is to directly use the pre-trained OrientCorrect from W-HMR to predict the body rotation in the world space and then combine it with the pose and shape output from HMR to generate the final human motion. It can be seen that it brings nearly 14\% improvement. From Tab.~\ref{tab:global}, it can be seen that using the camera rotations output from SPEC's CamCalib to correct brings only about 10\% boost.

Moreover, the most impressive is the second porting me-thod. We utilize the features and body orientation output by HMR to replace the original input part of OrientCorrect. In other words, we train an OrientCorrect adapted to HMR. The effect is pronounced, and the reconstruction accuracy is again improved by one grade. To avoid misunderstandings, you may think OrientCorrect is not as good as CamCalib or SPEC because the results in Tab.~\ref{tab:orientcorrect} are not as good as those of HMR and SPEC in Tab.~\ref{tab:global}. This is because we are based on different baselines. We use the pre-trained weights from SPIN directly and do not do any tuning, so the final results should be worse. 

\begin{table}[h]

\caption{Ablation study results on SPEC-MTP. The weights are from SPIN and are not fine-tuned. $^\dagger$ indicates using orientation output by OrientCorrect of pre-trained W-HMR. $^*$ indicates HMR with an adaptive OrientCorrect module.}
\label{tab:orientcorrect}
\scriptsize%
\centering%
\setlength{\tabcolsep}{5.6mm}{
\begin{tabular}{clccc}
\hline
\multicolumn{2}{c}{Model}                       & W-MPJPE$\downarrow$        & PA-MPJPE$\downarrow$      & W-PVE$\downarrow$    \\ \hline
\multicolumn{2}{c}{HMR}                    &173.6     &\textbf{75.4}           &206.6    \\ \hline
\multicolumn{2}{c}{HMR$^\dagger$}                 &149.6     &\textbf{75.4}           &179.4     \\
\multicolumn{2}{c}{HMR$^*$}                 &\textbf{133.9}  &\textbf{75.4}  &\textbf{157.9}   \\ \hline
\end{tabular}}
\end{table}

\textbf{Design Details} To find the optimal design for our model, we conduct further ablation experiments, as shown in Tab.~\ref{tab:detail}. Firstly, we test different downsampling methods. Using markers for downsampling leads to better learning of human body features. Here, “markers" refer to mocap markers. The improved performance might be because mocap markers are a better proxy for inferring human body information~\citep{zhang2021we}. However, since using mocap markers conflicts with the vertex regressor, we only employ markers in the pure parameter regression.

\begin{table}[th]

\caption{Ablation experiments regarding the design details of modules. The first comparison involves vertex selection for downsampling. The second comparison examines the impact of different information inputs. The third comparison evaluates different designs of the vert graphomer. The fourth comparison evaluates different combinations of regressors. Within each comparison, all settings remain consistent except for the variations in module design.}
\label{tab:detail}
\scriptsize%
\centering%
\setlength{\tabcolsep}{2.5mm}{
\begin{tabular}{ccccc}
\hline
\multirow{2}{*}{Design Details} & \multicolumn{4}{c}{AGORA}                                     \\ \cline{2-5} 
                                & NMVE$\downarrow$          & NMJE$\downarrow$          & MVE$\downarrow$           & MPJPE$\downarrow$         \\ \hline
sparse vertices                 & 88.5          & 94.0          & 76.1          & 80.8          \\
markers                          & \textbf{84.7} & \textbf{89.7} & \textbf{72.8} & \textbf{77.1} \\ \hline
3D bbox info                    & 82.3          & 88.5          & 74.9          & 80.5          \\
5D glob info                    & \textbf{80.7} & \textbf{87.7} & \textbf{73.4} & \textbf{79.8} \\ \hline
vert graphomer(indirect)        & 83.0          & 90.2          & 76.4          & 83.0          \\
vert graphomer(direct)          & \textbf{82.4} & \textbf{88.6} & \textbf{75.8} & \textbf{81.5} \\ \hline
1 param + 2 vert                & \textbf{82.0}   & 91.0            & \textbf{70.5} & 78.3          \\
2 param + 1 vert                & 82.9          & \textbf{89.5} & 71.0            & \textbf{77.0}   \\ \hline
\end{tabular}
}
\end{table}

We also compare our 5D global information with the 3D bounding box information proposed by CLIFF~\citep{cliff}. It is clear that, apart from the relative position information provided by the bounding box, the original image size also contains essential information for our task.

Additionally, we design different vertex regressors. One approach involves adding the output values to the input vertex coordinates to obtain the final results (indirect), while the other directly outputs the 3D vertex coordinates (direct). Evaluation results demonstrate that the direct approach is more efficient and effective.

Lastly, we test various combinations of the hybrid regression module. Both approaches show advantages, but it is apparent that the combination of “2 param$+$1 vert"  has clear benefits regarding model size and inference efficiency. Therefore, we ultimately select this combination as our final regression module.

\begin{figure}[h]
    \centering
    \includegraphics[width=0.99\linewidth]{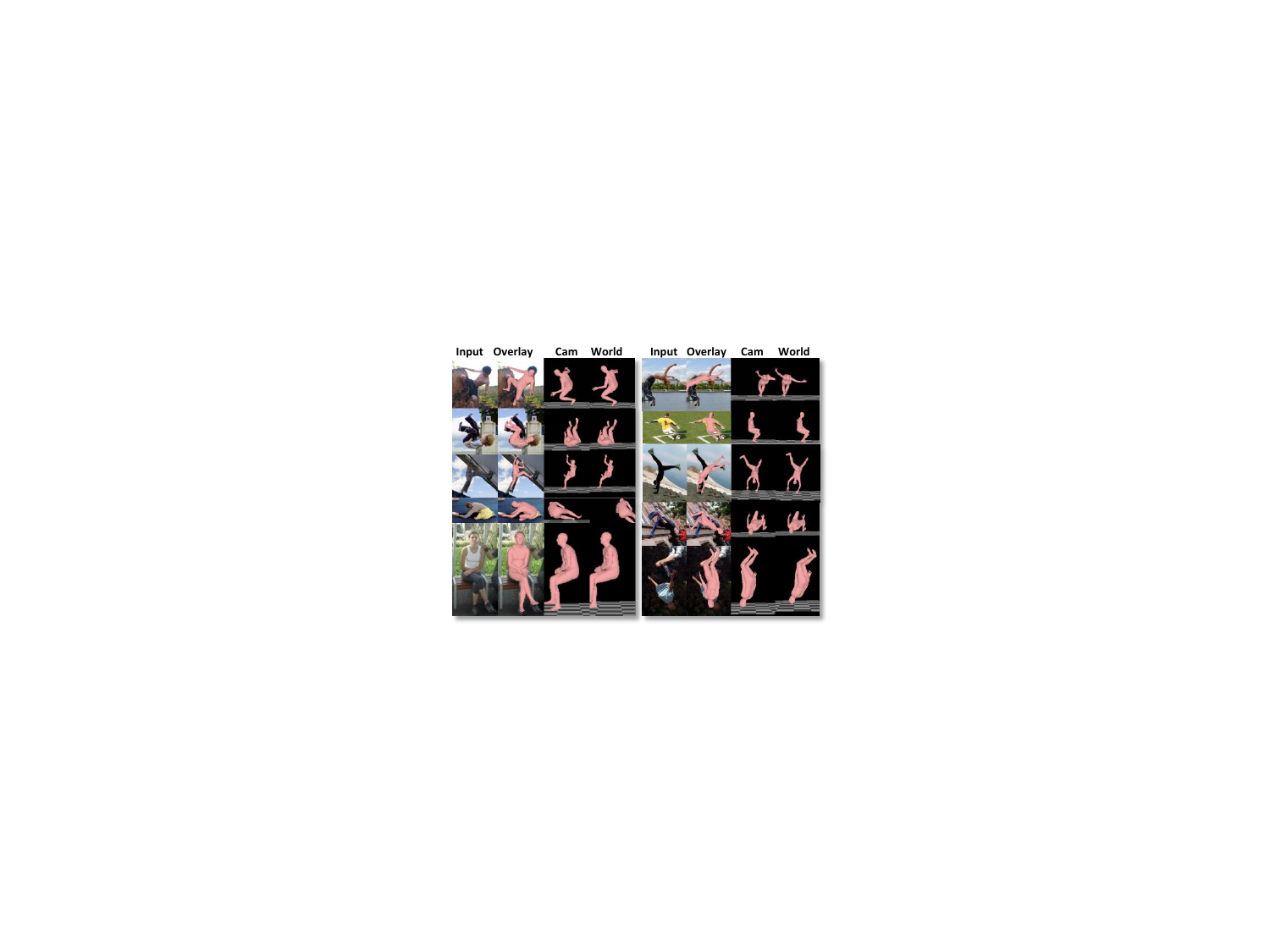}
    \caption{Failure cases on unusual samples. Although our model performs well in most cases, W-HMR still fails in some extreme situations. For example, humans have difficult poses, body parts are hard to distinguish from the background, and the cross of limbs leads to mesh penetration.}
    \label{fig:fail}
\end{figure}

\section{Discussion}
{\textbf{Difference} It is not the first time that predicting the distance between a person and a camera has been proposed. To understand our innovation, we take \citep{moon2019camera} as an example to discuss the differences between our work. Different Methods: The neural network in our work directly predicts the person's distance from the camera. But they first calculate values $k$ based on the sizes of the bounding box and image. Then, they predict a correction factor $\gamma$ through a neural network and get the absolute depth value using $k$ and $\gamma$. Different Theories: Our work mainly relies on the degree of distortion of humans in the images. The model tends to predict shorter distances when humans are more distorted. But \citep{moon2019camera} mainly relies on the human size in images. A more miniature human means this human stands farther away from the camera. Different Purposes: We use predicted distance to calculate focal lengths. Our ultimate goal is to achieve a perspective camera model so that we can make full use of distorted images. As for \citep{moon2019camera}, they want to recover accurate relative positions for multi-person pose estimation.}

\textbf{Limitation} Of course, W-HMR is still not a perfect model, and for this reason, we show the failure cases in Fig~\ref{fig:fail}. Our model can still not achieve accurate reconstruction when facing extreme poses (yoga, parkour), which may be due to the lack of relevant training data or the iterative method's limitations. Secondly, when the human limbs are close to the body, there is often through-molding, which requires an extra optimization-based method to improve it or a human model with a higher degree of freedom. Third, W-HMR seems helpless against some severely distorted images taken at extremely rare angles (e.g., the image in the third row from the left). After all, as a weakly supervised method, W-HMR is limited in its ability to predict focal length without rich annotations for training. It is also true that W-HMR cannot correctly establish the relationship between the person and the ground for images lacking background references. These will be one of the directions worth researching in the future.

\textbf{Future} Another vital direction for human motion recovery in world space is to capture the global trajectory of 3D human bodies. W-HMR mainly solves the problems caused by camera position and rotation. Although W-HMR can recover the human body posture in world space, it cannot obtain the absolute spatial position of the human body. Absolute spatial position is essential for the practical application of motion capture. At present, most related research still needs to rely on additional equipment to assist. 
{Some methods that can predict 3D pose~\citep{moon2019camera,mehta2017vnect,shimada2020physcap} or SMPL~\citep{sun2022putting,tripathi20233d} in the world coordinate from monocular images have been proposed but still far from practical.}
This will be an important future research direction for motion capture based on monocular images.

\section{Conclusion}
We proposed the first model capable of reconstructing human bodies accurately and reasonably in both camera and world coordinates. To achieve this, we designed a weak-supervised camera calibration method. “Weak-Supervised" solves the problem of insufficient focal length labels. “Camera calibration" solves the problem caused by image distortion. The focal length predictor can generate appropriate focal lengths based on body distortion information, eliminating the dependence on background information and focal length labels, which are more suitable for human recovery tasks.

Additionally, we introduced a new orientation correction module that is simpler and more robust than previous methods. Moreover, we proved it can be ported to other models and work. We achieved stable training and impressive performance across multiple datasets in both coordinates with a carefully designed training paradigm and model framework. We hope that our work will contribute to the study of human motion recovery in world space. The codes and related materials will be kept updated. Everyone is welcome to acquire them from our project page \href{https://yw0208.github.io/w-hmr/}{\textcolor{magenta}{https://yw0208.github.io/w-hmr/}}.




\begin{thebibliography}{65}
\providecommand{\natexlab}[1]{#1}
\providecommand{\url}[1]{{#1}}
\providecommand{\urlprefix}{URL }
\expandafter\ifx\csname urlstyle\endcsname\relax
  \providecommand{\doi}[1]{DOI~\discretionary{}{}{}#1}\else
  \providecommand{\doi}{DOI~\discretionary{}{}{}\begingroup \urlstyle{rm}\Url}\fi
\providecommand{\eprint}[2][]{\url{#2}}

\bibitem[{Andriluka et~al.(2014)Andriluka, Pishchulin, Gehler, and Schiele}]{mpii}
Andriluka M, Pishchulin L, Gehler P, Schiele B (2014) 2d human pose estimation: New benchmark and state of the art analysis. In: Proceedings of the IEEE Conference on computer Vision and Pattern Recognition, pp 3686--3693

\bibitem[{Anguelov et~al.(2005)Anguelov, Srinivasan, Koller, Thrun, Rodgers, and Davis}]{anguelov2005scape}
Anguelov D, Srinivasan P, Koller D, Thrun S, Rodgers J, Davis J (2005) Scape: shape completion and animation of people. In: ACM SIGGRAPH 2005 Papers, pp 408--416

\bibitem[{Cai et~al.(2022)Cai, Ren, Zeng, Lin, Yu, Wang, Fan, Gao, Yu, Pan, Hong, Zhang, Loy, Yang, and Liu}]{cai2022humman}
Cai Z, Ren D, Zeng A, Lin Z, Yu T, Wang W, Fan X, Gao Y, Yu Y, Pan L, Hong F, Zhang M, Loy CC, Yang L, Liu Z (2022) {HuMMan}: Multi-modal 4d human dataset for versatile sensing and modeling. In: 17th European Conference on Computer Vision, Tel Aviv, Israel, October 23--27, 2022, Proceedings, Part VII, Springer, pp 557--577

\bibitem[{Cho et~al.(2022)Cho, Youwang, and Oh}]{fastmetro}
Cho J, Youwang K, Oh TH (2022) Cross-attention of disentangled modalities for 3d human mesh recovery with transformers. In: European Conference on Computer Vision, Springer, pp 342--359

\bibitem[{Choi et~al.(2020)Choi, Moon, and Lee}]{choi2020pose2mesh}
Choi H, Moon G, Lee KM (2020) Pose2mesh: Graph convolutional network for 3d human pose and mesh recovery from a 2d human pose. In: Computer Vision--ECCV 2020: 16th European Conference, Glasgow, UK, August 23--28, 2020, Proceedings, Part VII 16, Springer, pp 769--787

\bibitem[{Dosovitskiy et~al.(2020)Dosovitskiy, Beyer, Kolesnikov, Weissenborn, Zhai, Unterthiner, Dehghani, Minderer, Heigold, Gelly et~al.}]{vit}
Dosovitskiy A, Beyer L, Kolesnikov A, Weissenborn D, Zhai X, Unterthiner T, Dehghani M, Minderer M, Heigold G, Gelly S, et~al. (2020) An image is worth 16x16 words: Transformers for image recognition at scale. arXiv preprint arXiv:201011929

\bibitem[{Fang et~al.(2023)Fang, Chen, Fan, Shuai, Li, and Zhang}]{propose}
Fang Q, Chen K, Fan Y, Shuai Q, Li J, Zhang W (2023) Learning analytical posterior probability for human mesh recovery. In: Proceedings of the IEEE/CVF Conference on Computer Vision and Pattern Recognition, pp 8781--8791

\bibitem[{Goel et~al.(2023)Goel, Pavlakos, Rajasegaran, Kanazawa, and Malik}]{humans4d}
Goel S, Pavlakos G, Rajasegaran J, Kanazawa A, Malik J (2023) Humans in 4d: Reconstructing and tracking humans with transformers. arXiv preprint arXiv:230520091

\bibitem[{Gu et~al.(2018)Gu, Sun, Ross, Vondrick, Pantofaru, Li, Vijayanarasimhan, Toderici, Ricco, Sukthankar et~al.}]{ava}
Gu C, Sun C, Ross DA, Vondrick C, Pantofaru C, Li Y, Vijayanarasimhan S, Toderici G, Ricco S, Sukthankar R, et~al. (2018) Ava: A video dataset of spatio-temporally localized atomic visual actions. In: Proceedings of the IEEE conference on computer vision and pattern recognition, pp 6047--6056

\bibitem[{Guan et~al.(2009)Guan, Weiss, Balan, and Black}]{guan2009estimating}
Guan P, Weiss A, Balan AO, Black MJ (2009) Estimating human shape and pose from a single image. In: 2009 IEEE 12th International Conference on Computer Vision, IEEE, pp 1381--1388

\bibitem[{G{\"u}ler et~al.(2018)G{\"u}ler, Neverova, and Kokkinos}]{densepose}
G{\"u}ler RA, Neverova N, Kokkinos I (2018) Densepose: Dense human pose estimation in the wild. In: Proceedings of the IEEE conference on computer vision and pattern recognition, pp 7297--7306

\bibitem[{He et~al.(2016)He, Zhang, Ren, and Sun}]{resnet}
He K, Zhang X, Ren S, Sun J (2016) Deep residual learning for image recognition. In: Proceedings of the IEEE conference on computer vision and pattern recognition, pp 770--778

\bibitem[{Ionescu et~al.(2013)Ionescu, Papava, Olaru, and Sminchisescu}]{human36m}
Ionescu C, Papava D, Olaru V, Sminchisescu C (2013) Human3. 6m: Large scale datasets and predictive methods for 3d human sensing in natural environments. IEEE transactions on pattern analysis and machine intelligence 36(7):1325--1339

\bibitem[{Joo et~al.(2021)Joo, Neverova, and Vedaldi}]{eft}
Joo H, Neverova N, Vedaldi A (2021) Exemplar fine-tuning for 3d human model fitting towards in-the-wild 3d human pose estimation. In: 2021 International Conference on 3D Vision (3DV), IEEE, pp 42--52

\bibitem[{Kanazawa et~al.(2018)Kanazawa, Black, Jacobs, and Malik}]{hmr}
Kanazawa A, Black MJ, Jacobs DW, Malik J (2018) End-to-end recovery of human shape and pose. In: Proceedings of the IEEE conference on computer vision and pattern recognition, pp 7122--7131

\bibitem[{Kanazawa et~al.(2019)Kanazawa, Zhang, Felsen, and Malik}]{insta}
Kanazawa A, Zhang JY, Felsen P, Malik J (2019) Learning 3d human dynamics from video. In: Proceedings of the IEEE/CVF conference on computer vision and pattern recognition, pp 5614--5623

\bibitem[{Kissos et~al.(2020)Kissos, Fritz, Goldman, Meir, Oks, and Kliger}]{kissos2020beyond}
Kissos I, Fritz L, Goldman M, Meir O, Oks E, Kliger M (2020) Beyond weak perspective for monocular 3d human pose estimation. In: Computer Vision--ECCV 2020 Workshops: Glasgow, UK, August 23--28, 2020, Proceedings, Part II 16, Springer, pp 541--554

\bibitem[{Kocabas et~al.(2020)Kocabas, Athanasiou, and Black}]{vibe}
Kocabas M, Athanasiou N, Black MJ (2020) Vibe: Video inference for human body pose and shape estimation. In: Proceedings of the IEEE/CVF conference on computer vision and pattern recognition, pp 5253--5263

\bibitem[{Kocabas et~al.(2021{\natexlab{a}})Kocabas, Huang, Hilliges, and Black}]{kocabas2021pare}
Kocabas M, Huang CHP, Hilliges O, Black MJ (2021{\natexlab{a}}) Pare: Part attention regressor for 3d human body estimation. In: Proceedings of the IEEE/CVF International Conference on Computer Vision, pp 11127--11137

\bibitem[{Kocabas et~al.(2021{\natexlab{b}})Kocabas, Huang, Tesch, M{\"u}ller, Hilliges, and Black}]{spec}
Kocabas M, Huang CHP, Tesch J, M{\"u}ller L, Hilliges O, Black MJ (2021{\natexlab{b}}) Spec: Seeing people in the wild with an estimated camera. In: Proceedings of the IEEE/CVF International Conference on Computer Vision, pp 11035--11045

\bibitem[{Kolotouros et~al.(2019{\natexlab{a}})Kolotouros, Pavlakos, Black, and Daniilidis}]{spin}
Kolotouros N, Pavlakos G, Black MJ, Daniilidis K (2019{\natexlab{a}}) Learning to reconstruct 3d human pose and shape via model-fitting in the loop. In: Proceedings of the IEEE/CVF international conference on computer vision, pp 2252--2261

\bibitem[{Kolotouros et~al.(2019{\natexlab{b}})Kolotouros, Pavlakos, and Daniilidis}]{graphcmr}
Kolotouros N, Pavlakos G, Daniilidis K (2019{\natexlab{b}}) Convolutional mesh regression for single-image human shape reconstruction. In: Proceedings of the IEEE/CVF Conference on Computer Vision and Pattern Recognition, pp 4501--4510

\bibitem[{Kolotouros et~al.(2021)Kolotouros, Pavlakos, Jayaraman, and Daniilidis}]{prohmr}
Kolotouros N, Pavlakos G, Jayaraman D, Daniilidis K (2021) Probabilistic modeling for human mesh recovery. In: Proceedings of the IEEE/CVF international conference on computer vision, pp 11605--11614

\bibitem[{Li et~al.(2021)Li, Xu, Chen, Bian, Yang, and Lu}]{li2021hybrik}
Li J, Xu C, Chen Z, Bian S, Yang L, Lu C (2021) Hybrik: A hybrid analytical-neural inverse kinematics solution for 3d human pose and shape estimation. In: Proceedings of the IEEE/CVF conference on computer vision and pattern recognition, pp 3383--3393

\bibitem[{Li et~al.(2023)Li, Liu, Lai, and Yang}]{li2023mili}
Li K, Liu Y, Lai YK, Yang J (2023) Mili: Multi-person inference from a low-resolution image. Fundamental Research 3(3):434--441

\bibitem[{Li et~al.(2022{\natexlab{a}})Li, Mao, Girshick, and He}]{vitdet}
Li Y, Mao H, Girshick R, He K (2022{\natexlab{a}}) Exploring plain vision transformer backbones for object detection. In: European Conference on Computer Vision, Springer, pp 280--296

\bibitem[{Li et~al.(2022{\natexlab{b}})Li, Liu, Zhang, Xu, and Yan}]{cliff}
Li Z, Liu J, Zhang Z, Xu S, Yan Y (2022{\natexlab{b}}) Cliff: Carrying location information in full frames into human pose and shape estimation. In: European Conference on Computer Vision, Springer, pp 590--606

\bibitem[{Lin et~al.(2021{\natexlab{a}})Lin, Wang, and Liu}]{metro}
Lin K, Wang L, Liu Z (2021{\natexlab{a}}) End-to-end human pose and mesh reconstruction with transformers. In: Proceedings of the IEEE/CVF conference on computer vision and pattern recognition, pp 1954--1963

\bibitem[{Lin et~al.(2021{\natexlab{b}})Lin, Wang, and Liu}]{lin2021mesh}
Lin K, Wang L, Liu Z (2021{\natexlab{b}}) Mesh graphormer. In: Proceedings of the IEEE/CVF International Conference on Computer Vision, pp 12939--12948

\bibitem[{Lin et~al.(2014)Lin, Maire, Belongie, Hays, Perona, Ramanan, Doll{\'a}r, and Zitnick}]{coco}
Lin TY, Maire M, Belongie S, Hays J, Perona P, Ramanan D, Doll{\'a}r P, Zitnick CL (2014) Microsoft coco: Common objects in context. In: Computer Vision--ECCV 2014: 13th European Conference, Zurich, Switzerland, September 6-12, 2014, Proceedings, Part V 13, Springer, pp 740--755

\bibitem[{Loper et~al.(2015)Loper, Mahmood, Romero, Pons-Moll, and Black}]{loper2015smpl}
Loper M, Mahmood N, Romero J, Pons-Moll G, Black MJ (2015) Smpl: A skinned multi-person linear model. ACM transactions on graphics (TOG) 34(6):1--16

\bibitem[{Mehta et~al.(2017{\natexlab{a}})Mehta, Rhodin, Casas, Fua, Sotnychenko, Xu, and Theobalt}]{MPII3d}
Mehta D, Rhodin H, Casas D, Fua P, Sotnychenko O, Xu W, Theobalt C (2017{\natexlab{a}}) Monocular 3d human pose estimation in the wild using improved cnn supervision. In: 2017 international conference on 3D vision (3DV), IEEE, pp 506--516

\bibitem[{Mehta et~al.(2017{\natexlab{b}})Mehta, Sridhar, Sotnychenko, Rhodin, Shafiei, Seidel, Xu, Casas, and Theobalt}]{mehta2017vnect}
Mehta D, Sridhar S, Sotnychenko O, Rhodin H, Shafiei M, Seidel HP, Xu W, Casas D, Theobalt C (2017{\natexlab{b}}) Vnect: Real-time 3d human pose estimation with a single rgb camera. Acm transactions on graphics (tog) 36(4):1--14

\bibitem[{Moon and Lee(2020)}]{I2l-meshnet}
Moon G, Lee KM (2020) I2l-meshnet: Image-to-lixel prediction network for accurate 3d human pose and mesh estimation from a single rgb image. In: Computer Vision--ECCV 2020: 16th European Conference, Glasgow, UK, August 23--28, 2020, Proceedings, Part VII 16, Springer, pp 752--768

\bibitem[{Moon et~al.(2019)Moon, Chang, and Lee}]{moon2019camera}
Moon G, Chang JY, Lee KM (2019) Camera distance-aware top-down approach for 3d multi-person pose estimation from a single rgb image. In: Proceedings of the IEEE/CVF international conference on computer vision, pp 10133--10142

\bibitem[{Moon et~al.(2022)Moon, Choi, and Lee}]{hand4whole}
Moon G, Choi H, Lee KM (2022) Accurate 3d hand pose estimation for whole-body 3d human mesh estimation. In: Proceedings of the IEEE/CVF Conference on Computer Vision and Pattern Recognition, pp 2308--2317

\bibitem[{Patel et~al.(2021)Patel, Huang, Tesch, Hoffmann, Tripathi, and Black}]{agora}
Patel P, Huang CHP, Tesch J, Hoffmann DT, Tripathi S, Black MJ (2021) {AGORA}: Avatars in geography optimized for regression analysis. In: Proceedings IEEE/CVF Conf.~on Computer Vision and Pattern Recognition ({CVPR})

\bibitem[{Pavlakos et~al.(2019)Pavlakos, Choutas, Ghorbani, Bolkart, Osman, Tzionas, and Black}]{smplx}
Pavlakos G, Choutas V, Ghorbani N, Bolkart T, Osman AA, Tzionas D, Black MJ (2019) Expressive body capture: 3d hands, face, and body from a single image. In: Proceedings of the IEEE/CVF conference on computer vision and pattern recognition, pp 10975--10985

\bibitem[{Pons-Moll et~al.(2015)Pons-Moll, Romero, Mahmood, and Black}]{pons2015dyna}
Pons-Moll G, Romero J, Mahmood N, Black MJ (2015) Dyna: A model of dynamic human shape in motion. ACM Transactions on Graphics (TOG) 34(4):1--14

\bibitem[{Shetty et~al.(2023)Shetty, Birkhold, Jaganathan, Strobel, Kowarschik, Maier, and Egger}]{shetty2023pliks}
Shetty K, Birkhold A, Jaganathan S, Strobel N, Kowarschik M, Maier A, Egger B (2023) Pliks: A pseudo-linear inverse kinematic solver for 3d human body estimation. In: Proceedings of the IEEE/CVF Conference on Computer Vision and Pattern Recognition, pp 574--584

\bibitem[{Shimada et~al.(2020)Shimada, Golyanik, Xu, and Theobalt}]{shimada2020physcap}
Shimada S, Golyanik V, Xu W, Theobalt C (2020) Physcap: Physically plausible monocular 3d motion capture in real time. ACM Transactions on Graphics (ToG) 39(6):1--16

\bibitem[{Shin et~al.(2024)Shin, Kim, Halilaj, and Black}]{shin2024wham}
Shin S, Kim J, Halilaj E, Black MJ (2024) Wham: Reconstructing world-grounded humans with accurate 3d motion. In: Proceedings of the IEEE/CVF Conference on Computer Vision and Pattern Recognition, pp 2070--2080

\bibitem[{Sun et~al.(2022)Sun, Liu, Bao, Fu, Mei, and Black}]{sun2022putting}
Sun Y, Liu W, Bao Q, Fu Y, Mei T, Black MJ (2022) Putting people in their place: Monocular regression of 3d people in depth. In: Proceedings of the IEEE/CVF Conference on Computer Vision and Pattern Recognition, pp 13243--13252

\bibitem[{Tian et~al.(2023)Tian, Zhang, Liu, and Wang}]{tian2022survey}
Tian Y, Zhang H, Liu Y, Wang L (2023) Recovering 3d human mesh from monocular images: A survey. IEEE transactions on pattern analysis and machine intelligence

\bibitem[{Tripathi et~al.(2023)Tripathi, M{\"u}ller, Huang, Taheri, Black, and Tzionas}]{tripathi20233d}
Tripathi S, M{\"u}ller L, Huang CHP, Taheri O, Black MJ, Tzionas D (2023) 3d human pose estimation via intuitive physics. In: Proceedings of the IEEE/CVF conference on computer vision and pattern recognition, pp 4713--4725

\bibitem[{Varol et~al.(2018)Varol, Ceylan, Russell, Yang, Yumer, Laptev, and Schmid}]{varol2018bodynet}
Varol G, Ceylan D, Russell B, Yang J, Yumer E, Laptev I, Schmid C (2018) Bodynet: Volumetric inference of 3d human body shapes. In: Proceedings of the European conference on computer vision (ECCV), pp 20--36

\bibitem[{Von~Marcard et~al.(2018)Von~Marcard, Henschel, Black, Rosenhahn, and Pons-Moll}]{3dpw}
Von~Marcard T, Henschel R, Black MJ, Rosenhahn B, Pons-Moll G (2018) Recovering accurate 3d human pose in the wild using imus and a moving camera. In: Proceedings of the European Conference on Computer Vision (ECCV), pp 601--617

\bibitem[{Wang et~al.(2023)Wang, Ge, Mei, Cai, Sun, Wang, Shen, Yang, and Komura}]{wang2023zolly}
Wang W, Ge Y, Mei H, Cai Z, Sun Q, Wang Y, Shen C, Yang L, Komura T (2023) Zolly: Zoom focal length correctly for perspective-distorted human mesh reconstruction. In: Proceedings of the IEEE/CVF International Conference on Computer Vision, pp 3925--3935

\bibitem[{Wang and Daniilidis(2023)}]{wang2023refit}
Wang Y, Daniilidis K (2023) Refit: Recurrent fitting network for 3d human recovery. In: Proceedings of the IEEE/CVF International Conference on Computer Vision, pp 14644--14654

\bibitem[{Wei et~al.(2022)Wei, Lin, Liu, and Liao}]{mpsnet}
Wei WL, Lin JC, Liu TL, Liao HYM (2022) Capturing humans in motion: Temporal-attentive 3d human pose and shape estimation from monocular video. In: Proceedings of the IEEE/CVF Conference on Computer Vision and Pattern Recognition, pp 13211--13220

\bibitem[{Wu et~al.(2019)Wu, Zheng, Zhao, Li, Yan, Liang, Wang, Zhou, Lin, Fu et~al.}]{aic}
Wu J, Zheng H, Zhao B, Li Y, Yan B, Liang R, Wang W, Zhou S, Lin G, Fu Y, et~al. (2019) Large-scale datasets for going deeper in image understanding. In: 2019 IEEE International Conference on Multimedia and Expo (ICME), IEEE, pp 1480--1485

\bibitem[{Xu et~al.(2020)Xu, Bazavan, Zanfir, Freeman, Sukthankar, and Sminchisescu}]{xu2020ghum}
Xu H, Bazavan EG, Zanfir A, Freeman WT, Sukthankar R, Sminchisescu C (2020) Ghum \& ghuml: Generative 3d human shape and articulated pose models. In: Proceedings of the IEEE/CVF Conference on Computer Vision and Pattern Recognition, pp 6184--6193

\bibitem[{Xu et~al.(2022{\natexlab{a}})Xu, Zhang, Zhang, and Tao}]{xu2022vitpose}
Xu Y, Zhang J, Zhang Q, Tao D (2022{\natexlab{a}}) Vitpose: Simple vision transformer baselines for human pose estimation. Advances in Neural Information Processing Systems 35:38571--38584

\bibitem[{Xu et~al.(2022{\natexlab{b}})Xu, Zhang, Zhang, and Tao}]{xu2022vitpose+}
Xu Y, Zhang J, Zhang Q, Tao D (2022{\natexlab{b}}) Vitpose+: Vision transformer foundation model for generic body pose estimation. arXiv preprint arXiv:221204246

\bibitem[{Zanfir et~al.(2021{\natexlab{a}})Zanfir, Bazavan, Zanfir, Freeman, Sukthankar, and Sminchisescu}]{hund}
Zanfir A, Bazavan EG, Zanfir M, Freeman WT, Sukthankar R, Sminchisescu C (2021{\natexlab{a}}) Neural descent for visual 3d human pose and shape. In: Proceedings of the IEEE/CVF Conference on Computer Vision and Pattern Recognition, pp 14484--14493

\bibitem[{Zanfir et~al.(2021{\natexlab{b}})Zanfir, Zanfir, Bazavan, Freeman, Sukthankar, and Sminchisescu}]{thundr}
Zanfir M, Zanfir A, Bazavan EG, Freeman WT, Sukthankar R, Sminchisescu C (2021{\natexlab{b}}) Thundr: Transformer-based 3d human reconstruction with markers. In: Proceedings of the IEEE/CVF International Conference on Computer Vision, pp 12971--12980

\bibitem[{Zhang et~al.(2020{\natexlab{a}})Zhang, Cao, Lu, Ouyang, and Sun}]{danet}
Zhang H, Cao J, Lu G, Ouyang W, Sun Z (2020{\natexlab{a}}) Learning 3d human shape and pose from dense body parts. IEEE Transactions on Pattern Analysis and Machine Intelligence

\bibitem[{Zhang et~al.(2021{\natexlab{a}})Zhang, Tian, Zhou, Ouyang, Liu, Wang, and Sun}]{pymaf}
Zhang H, Tian Y, Zhou X, Ouyang W, Liu Y, Wang L, Sun Z (2021{\natexlab{a}}) Pymaf: 3d human pose and shape regression with pyramidal mesh alignment feedback loop. In: Proceedings of the IEEE/CVF International Conference on Computer Vision, pp 11446--11456

\bibitem[{Zhang et~al.(2023{\natexlab{a}})Zhang, Tian, Zhang, Li, An, Sun, and Liu}]{pymaf-x}
Zhang H, Tian Y, Zhang Y, Li M, An L, Sun Z, Liu Y (2023{\natexlab{a}}) Pymaf-x: Towards well-aligned full-body model regression from monocular images. IEEE Transactions on Pattern Analysis and Machine Intelligence

\bibitem[{Zhang et~al.(2020{\natexlab{b}})Zhang, Huang, and Wang}]{zhang2020object}
Zhang T, Huang B, Wang Y (2020{\natexlab{b}}) Object-occluded human shape and pose estimation from a single color image. In: Proceedings of the IEEE/CVF conference on computer vision and pattern recognition, pp 7376--7385

\bibitem[{Zhang et~al.(2021{\natexlab{b}})Zhang, Black, and Tang}]{zhang2021we}
Zhang Y, Black MJ, Tang S (2021{\natexlab{b}}) We are more than our joints: Predicting how 3d bodies move. In: Proceedings of the IEEE/CVF Conference on Computer Vision and Pattern Recognition, pp 3372--3382

\bibitem[{Zhang et~al.(2023{\natexlab{b}})Zhang, Zhang, Hu, Yi, Zhang, and Liu}]{zhang2023real}
Zhang Y, Zhang H, Hu L, Yi H, Zhang S, Liu Y (2023{\natexlab{b}}) Real-time monocular full-body capture in world space via sequential proxy-to-motion learning. arXiv preprint arXiv:230701200

\bibitem[{Zheng et~al.(2019)Zheng, Yu, Wei, Dai, and Liu}]{zheng2019deephuman}
Zheng Z, Yu T, Wei Y, Dai Q, Liu Y (2019) Deephuman: 3d human reconstruction from a single image. In: Proceedings of the IEEE/CVF International Conference on Computer Vision, pp 7739--7749

\bibitem[{Zhou et~al.(2010)Zhou, Fu, Liu, Cohen-Or, and Han}]{zhou2010parametric}
Zhou S, Fu H, Liu L, Cohen-Or D, Han X (2010) Parametric reshaping of human bodies in images. ACM transactions on graphics (TOG) 29(4):1--10

\bibitem[{Zuffi and Black(2015)}]{zuffi2015stitched}
Zuffi S, Black MJ (2015) The stitched puppet: A graphical model of 3d human shape and pose. In: Proceedings of the IEEE Conference on Computer Vision and Pattern Recognition, pp 3537--3546

\end{thebibliography}





\section{About Datasets}
\label{sec:dataset}
In our experiments, we utilize the following datasets: Human 3.6M~\citep{human36m}, 3DPW~\citep{3dpw}, MPII~\citep{mpii}, COCO~\citep{coco}, MPI-INF-3D~\citep{MPII3d}, AVA
~\citep{ava}, AIC~\citep{aic}, Insta~\citep{insta}, AGORA~\citep{agora}, SPEC-SYN~\citep{spec}, SPEC-MTP~\citep{spec}, and HuMMan~\citep{cai2022humman}. To evaluate our model, we conduct experiments on the distorted datasets AGORA, HuMMan, and SPEC-MTP. Furthermore, following traditional practices\citep{mpsnet, pymaf}, we also evaluate our model on 3DPW, Human 3.6M, and MPII-INF-3D. We employ AGORA's evaluation platform as a standard for ablation experiments due to its diverse scenes, characters, and shooting methods. We believe that the evaluation results obtained on AGORA are more representative. For human recovery in world space, We additionally perform an ablation study on SPEC-SYN to validate the robustness of our method. Following are details about each dataset.

\textbf{3DPW}'~\citep{3dpw} full name is “3D Human in the Wild."  It is an outdoor dataset that captures actor information using motion capture devices. This dataset is known for its difficulty and has been widely used by researchers to demonstrate the robustness of their methods. Following past conventions, we train two models, one without 3DPW for training and another using 3DPW for training, to evaluate our model's robustness. 

\textbf{Human 3.6M}~\citep{human36m} is widely used as a standard dataset to assess the estimation of 3D human pose and the recovery of human motion. To reduce data redundancy, the original videos are down-sampled from 50 to 10 fps, resulting in 312,188 frames available for training and 26859 frames for evaluation. Following established protocols, we utilize five subjects (S1, S5, S6, S7, S8) for training and two subjects (S9, S11) for evaluation. 

\textbf{MPII}~\citep{mpii} is collected from YouTube and comprises various activities in various scenes. Based on~\citep{pymaf-x,humans4d}, we only utilize the portion with pseudo 3D labels, containing approximately 14,000 images. 

\textbf{COCO}~\citep{coco} is a large-scale image dataset widely used for various computer vision tasks such as object detection, image segmentation, and image classification. For our task, we primarily utilize the part including human subjects. COCO provides 2D joint location labels. Based on this, we use corresponding pseudo 3D labels generated by \citep{cliff,eft,prohmr}, following \citep{pymaf-x,cliff,humans4d}.

\textbf{MPI-INF-3D}~\citep{MPII3d,mehta2017vnect} is a large-scale 3D dataset that includes both indoor and outdoor scenes. To reduce data redundancy, we extract every 10th frame. The dataset is divided into a training set and a test set according to their original partitioning. The training set comprises 96,507 images, while the test set contains 2929 images. We use pseudo SMPL parameter labels generated by \citep{eft}.

\textbf{AVA}~\citep{ava}, \textbf{AI Challenger}\citep{aic} and \textbf{InstaVariety}~\citep{insta} are three huge wild datasets that lack real 3D labels. However, in the Human4D~\citep{humans4d}, researchers utilized VitPose~\citep{xu2022vitpose} to obtain 2D keypoints and VitDet~\citep{vitdet} to generate bounding boxes for human targets. Based on these 2D pseudo labels, they employed ProHMR~\citep{prohmr} to acquire pseudo SMPL labels. Subsequently, we further optimize these pseudo-labeled datasets by selecting high-quality samples for training our model.

\begin{table*}[th]
\caption{This table illustrates the changes in the size of input images and features when using different backbones (ResNet50 or VitPose-B) and employing different downsampling methods (sparse vertices or markers). As for what symbols mean, please refer to Sec.~\ref{sec:extractor}.}
\label{tab:implement}
\scriptsize%
\centering%
\setlength{\tabcolsep}{5.25mm}{
\begin{tabular}{ccccc}
\hline
                             & ResNet50 + vertices                            & ResNet50 + markers                            & VitPose-B + vertices                            & VitPose-B + markers                            \\ \hline
\rowcolor[HTML]{EFEFEF} 
$H_{c}\times W_{c}$                  & 224$\times$224                             & 224$\times$224                             & 256$\times$192                              & 256$\times$192                              \\
$\{H_i \times W_i\}_{i=1}^3$ & \{14$\times$14,28$\times$28,56$\times$56\} & \{14$\times$14,28$\times$28,56$\times$56\} & \{32$\times$24,64$\times$48,128$\times$96\} & \{32$\times$24,64$\times$48,128$\times$96\} \\
\rowcolor[HTML]{EFEFEF} 
$C_{i}^{fc}$                 & 5                                          & 32                                         & 5                                           & 32                                          \\
$H_g \times W_g$                         & 21$\times$21                               & 8$\times$8                                 & 24$\times$18                                & 9$\times$7                                  \\
\rowcolor[HTML]{EFEFEF} 
$\{C_{i}^{s}\}_{i=1}^3$      & \{2205,2155,2155\}                         & \{2048,2144,2144\}                         & \{2160,2155,2155\}                          & \{2016,2144,2144\}                          \\ \hline
\end{tabular}}
\end{table*}

\textbf{AGORA}~\citep{agora} is a multi-person synthetic dataset that has gained recognition recently due to its precise SMPL-X labels, complex characters and scenes, and fair evaluation methods. The dataset is available in two resolutions: 1280x720 and 3840x2160. We exclusively utilize the low-resolution set in our work. The train, validation, and test sets are used according to their original partitioning. The bounding boxes used in the test set are obtained using the 2D keypoints generated by Hand4Whole~\citep{hand4whole}.

\textbf{SPEC-SYN} and \textbf{SPEC-MTP} are datasets proposed in SPEC~\citep{spec}. Inspired by AGORA, SPEC-SYN is also a synthetic dataset with 3,783 images synthesized from five large-scale virtual scenes. It offers a relatively high diversity of focal lengths. On the other hand, SPEC-MTP is a real dataset with SMPL-X labels generated by multi-view fitting. It consists of 3,284 images, which we solely use for evaluating our model. The dataset is challenging because of its serious distortion and special shooting angle.

\textbf{HuMMan}\citep{cai2022humman} is a high-precision human dataset constructed indoors. It includes labels such as keypoints, point clouds, meshes, SMPL parameters, textures, and more. We only utilize its SMPL labels. We sample every five images to form a training set comprising approximately 84,000 images. Due to the short focal length during data collection, it contains a significant amount of distorted data. Therefore, we also use HuMMan as one of our evaluation datasets. The partitioning of the train and test sets follows the original division provided by the dataset creators.

\section{Network Details}
\textbf{Pixel-aligned Feature Extractor \label{sec:extractor}}
We follow the feature extraction method of \citep{pymaf} but use different sampling points and backbones.
When a cropped image $I\in \mathbb{R} ^{H_{c}\times W_{c}}$ is inputted into the encoder, it generates an initial feature map $
\varPhi _0\in \mathbb{R} ^{C\times H_{0}\times W_{0}}
$. Subsequently, this feature map $\varPhi _0$ undergoes further processing through a series of deconvolution layers to yield high-resolution features $\left\{ \varPhi _i\in \mathbb{R} ^{C_i\times H_i\times W_i} \right\} _{i=1}^{3}$. For the first high-resolution feature map $\varPhi _1$, a $H_g \times W_g$ grid is defined to sample finer point-wise features. But, for the second and third feature maps, we project vertices of the mesh obtained from the previous step onto the feature map to sample point-wise features. In summary, we extract spatial features based on the following formula.

\begin{equation}
    \phi _{i}^{s}=\oplus \left( fc\left( bs\left( \varPhi _i,P_{i-1} \right) \right) \right),\, i\in\{1,\,2,\,3\}
\end{equation}
where $P_i$ represents the projected 2D point for sampling, $bs(\cdot)$ denotes bilinear sampling, and $fc(\cdot)$ refers to FC layers for reducing channel size. Finally, $\oplus$ signifies the concatenation of individual point-wise features $\left\{ \phi _{i,n}^{p,fc}\in \mathbb{R} ^{C_{i}^{fc}} \right\} _{n=1}^{N_i}$ into a one-dimensional feature $\phi _{i}^{s}\in \mathbb{R} ^{C_{i}^{s}}$, where $C_{i}^{s}=N_i \times C_{i}^{fc}$. $fc$ indicates the point-wise features has been through $fc(\cdot)$ for channel reduction and $N_i$ is the number of points for downsampling.

\textbf{Hybrid Regression Module \label{sec:regression}} For the spatial features $\{ \phi _{i}^{s} \}_{i=1}^{3}$, we have corresponding regression models. We utilize parameter regressors for the first two spatial features to obtain the initial motion. These regressors take spatial features $\phi _{i}^{s}$, global information $glob\_info$, and former SMPL parameters $\varTheta _{i-1}$ as inputs and output SMPL parameters $\varTheta _{i}$. The parameter regressors consist of several FC layers and iteratively produce the final result like
\begin{equation}
    \varTheta _i=\varTheta _{i-1}+FC\left( \oplus \left( \phi _{i}^{s},\,\,glob\_info,\ \varTheta _{i-1} \right) \right) 
\end{equation}
Note $\varTheta _0$ is set as the mean $\overline{\varTheta}$ following \citep{hmr}. 
As for the final regressor, we have two versions in total. The first version still involves the parameter regressor. The second version utilizes a vertex regressor to obtain a finer human body motion. After obtaining a coarse human motion using the first two regressors, we introduce a model-free vertex regressor based on the transformer. This vertex regressor takes spatial features $\phi _{3}^{s}$ and point-wise features $\left\{ \phi _{3,n}^{p}\in \mathbb{R} ^{C_{i}} \right\} _{n=1}^{N_i}$ as input. It's important to note that no “$fc$" exists in this case, indicating that these features do not undergo channel reduction. To reduce network complexity and adapt to the SMPL model, instead of directly recovering the 6890 vertices defined by SMPL, we recover 431 sparse vertices and gradually restore them to 6890 vertices using two linear layers.

\end{document}